\documentclass{article} 
\usepackage{iclr2025_conference,times}

\usepackage{amsmath,amsfonts,bm}

\def\eqref#1{equation~\ref{#1}}

\def\1{\bm{1}}

\DeclareMathAlphabet{\mathsfit}{\encodingdefault}{\sfdefault}{m}{sl}
\SetMathAlphabet{\mathsfit}{bold}{\encodingdefault}{\sfdefault}{bx}{n}

\usepackage{hyperref}
\usepackage{url}
\usepackage{graphicx}
\usepackage{wrapfig}
\usepackage[table]{xcolor}
\usepackage{amsthm}
\usepackage{algorithm}
\usepackage{algorithmic}
\title{Just Align $\bm{x}$: Aligning Predictions, Not Representations}

\author{Yuyao Zhang \quad Yuwei Hu \quad Ziyang Mai \quad Yu-Wing Tai \\
Dartmouth College
}

\iclrfinalcopy 

\newcommand{\ourwork}{\textbf{JA$\bm{x}$}~}
\newtheorem{proposition}{Proposition}
\definecolor{tableheader}{HTML}{E8EDF3}
\definecolor{tablegroup}{HTML}{F1F3F5}
\definecolor{tableresult}{HTML}{EEF5FC}
\definecolor{tableborder}{HTML}{A6AFBA}

\begin{document}

\maketitle

\begin{abstract}
Representation alignment has become an effective way to accelerate diffusion training, but its benefits do not transfer reliably to pixel-space clean-image prediction. In JiT, we find that auxiliary feature alignment can improve access to semantic features while reducing access to image variation needed for clean-image prediction, creating a mismatch between the auxiliary objective and the denoising task. This suggests a different principle: \textit{auxiliary supervision should improve the prediction target itself rather than impose a separate representation target}. We introduce ~\ourwork (Just Align $\bm{x}$), a prediction-supervision method that aligns clean-image predictions across noise levels. ~\ourwork couples a noisier student observation with a cleaner observation through a Markov degradation that preserves the original JiT input distribution. Under this coupling, the oracle prediction from the cleaner state has the same conditional mean as the optimal JiT target, while its conditional target covariance is no greater. Thus, oracle prediction alignment preserves the population JiT objective up to a constant while providing a lower-variance training target. To make this construction practical with an imperfect EMA teacher, ~\ourwork combines ground-truth supervision with a reliability-gated coupling band that selects nearby teacher states based on prediction risk. On ImageNet 256×256, ~\ourwork consistently improves FID and accelerates convergence across JiT-B/16, L/16, and H/16, without an external encoder or changes to the architecture or sampling procedure. Gradient diagnostics further show reduced minibatch gradient variance, while ablations demonstrate that the gains cannot be explained by time reweighting alone. These results show that prediction-space supervision provides a simple and principled alternative to representation alignment for pixel-space generative models.
\end{abstract}

\section{Introduction}

Diffusion and flow-based models have emerged as dominant paradigms for visual generation, transforming noise into data through iterative denoising~\citep{ho2020denoising} or continuous transport~\citep{song2020score, lipman2022flow, liu2022flow}. Diffusion Transformers (DiTs) provide a scalable architecture for modeling these generative dynamics~\citep{peebles2023scalable}. Beyond architectural scaling, recent studies~\citep{yu2024representation, singh2025matters, jiang2025no, wang2026sra, chefer2026self, wang2025diffuse} have shown that aligning or regularizing internal representations can substantially accelerate DiT training. However, most of this progress has been developed in latent diffusion~\citep{rombach2022high}, where an autoencoder compresses images before generative modeling. This introduces a reconstruction--generation trade-off: latent representations must preserve image details while remaining suitable for generation~\citep{leng2025repa, yao2025reconstruction, zheng2026diffusion, singh2026improved}. 

Pixel-space generation removes this intermediate representation but requires the model to directly predict high-dimensional images. Motivated by the low-dimensional manifold structure of natural images, JiT~\citep{li2026back} addresses this challenge through clean-image prediction with large pixel patches, providing a simple and effective framework without pretrained tokenizers or auxiliary representation objectives. This difference in prediction space raises a fundamental question: \textit{what auxiliary supervision is appropriate for pixel-space $\bm{x}$-prediction?}

A direct transfer of representation alignment is not always effective. PixelREPA~\citep{shin2026representation} shows that applying REPA can accelerate early learning but may degrade generation quality later in training, motivating adaptations to the mismatch between semantic features and pixel-level reconstruction. In the JiT settings we evaluate, transferred hidden self-alignment and regularization methods provide limited gains, while the advantage of REPA in low-resolution image generation can also reverse as model size decreases. One important difference is the input dimensionality: DiT-B/2 maps a $16$-dimensional latent patch to a $768$-dimensional hidden token, whereas JiT-B/16 maps $768$ pixel values to the same hidden width. Although these dimensions alone do not determine model capacity, they highlight a potential conflict between auxiliary feature learning and clean-image prediction. Our probes show increased linear accessibility of aligned semantic features together with reduced accessibility of clean-image variation not linearly explained by those features (Section~\ref{sec:motivation}). These observations motivate a different principle: rather than imposing an auxiliary representation target, auxiliary supervision should directly support the prediction target of the generative objective.

We therefore ask whether an auxiliary prediction target can preserve the conditional mean required by JiT while reducing the conditional target variance. Based on this principle, we introduce \ourwork (Just Align $\bm{x}$), which aligns clean-image predictions across noise levels. Specifically, we couple a cleaner observation with the student's noisier input through a Markov degradation that preserves the original student-input marginal. Under this coupling, the optimal clean-image predictions satisfy a martingale consistency property: averaging the cleaner prediction conditioned on the noisier observation recovers the optimal prediction at that noise level. With the original JiT time distribution and loss weighting, this oracle supervision induces the same population regression objective up to a parameter-independent constant, while its conditional target covariance is no greater than that of direct clean-image supervision.

In practice, \ourwork uses an exponential-moving-average (EMA) teacher to predict $\bm{x}$ from the cleaner observation. Because teacher errors can bias the auxiliary target, we retain supervision from the ground-truth clean image and introduce a reliability-gated coupling band that selects nearby teacher states according to an empirical prediction-risk criterion. This selection balances the potential variance reduction from cleaner targets against the error introduced by the evolving teacher. \ourwork requires no external encoder and leaves the JiT architecture and sampling procedure unchanged. Across model scales, it improves generation quality and accelerates convergence in training epochs. 

Our contributions are summarized as follows:
\begin{itemize}
    \item We empirically examine the limitations of hidden representation alignment in JiT and identify conditional target consistency as a guiding principle for auxiliary prediction supervision.
    \item We introduce Markov-coupled prediction supervision and establish conditional-mean consistency, population loss equivalence, and conditional covariance reduction for oracle predictions.
    \item We develop reliability-gated pair selection for an evolving EMA teacher and demonstrate improved generation quality and faster convergence across JiT model scales.
\end{itemize}
\section{Related Works}
\paragraph{Pixel-space diffusion}
Early diffusion models directly learn the generative process in RGB space~\citep{ho2020denoising,song2020score}, whereas modern large-scale systems commonly adopt latent diffusion to reduce the spatial and computational burden through a pretrained autoencoder~\citep{rombach2022high}. This efficiency, however, makes generation dependent on a separately learned latent space and its reconstruction properties. Recent works~\citep{baade2026latent, chen2025pixelflow, chen2026asymmetric, yu2026pixeldit, ma2026deco, li2026back} revisit end-to-end pixel-space generation from several directions: PixelFlow progressively increases spatial resolution along the generative trajectory~\citep{chen2025pixelflow}; PixelDiT and DeCo separate semantic or low-frequency modeling from pixel-level details~\citep{yu2026pixeldit, ma2026deco}; JiT instead studies a particularly minimal formulation based on large pixel patches and direct clean-image prediction, showing that high-dimensional RGB observations can be modeled without a tokenizer or an auxiliary decoder~\citep{li2026back}. Our work retains this minimal JiT architecture and focuses on accelerating its optimization without introducing a separate representation space or modifying its sampling process.
\paragraph{Representation Learning.} 
Beyond serving as a compressed generation space~\citep{rombach2022high}, learned representations can provide useful structure and supervision for generative modeling. Semantic tokenizers and representation autoencoders explore generation spaces that combine semantic information with image reconstruction~\citep{chen2025masked,zheng2026diffusion,singh2026improved}, while representation alignment methods use learned features to guide the optimization of generative models. REPA~\citep{yu2024representation} aligns intermediate diffusion features with representations from pretrained visual encoders, improving the hidden representation leading to better generation quality and training efficiency in latent DiTs. Subsequent work explores the choice of alignment targets~\citep{singh2025matters}, the placement and duration of supervision~\citep{liu2025efficient}, and the joint adaptation of the tokenizer and denoising network~\citep{leng2025repa}. Self-alignment methods~\citep{jiang2025no, wang2026sra, chefer2026self} further investigate whether useful representation guidance can be obtained from the generative model itself, reducing reliance on external encoders. However, changing the source of supervision does not necessarily remove the additional representational demands imposed by hidden-feature alignment. PixelREPA adapts feature alignment to pixel-space clean-image prediction~\citep{shin2026representation}. Nevertheless, it retains hidden representations as the alignment target, potentially imposing demands that compete with pixel reconstruction within shared model capacity. This motivates aligning clean-image predictions themselves: our method provides additional supervision for the denoising task without requiring hidden features to match a separate representation target, with conditional consistency and variance reduction established under a Markov coupling for optimal predictions.


\section{Motivation and Analysis}
\label{sec:motivation}

\paragraph{Preliminaries: \(x\)-prediction.}
Let \(X\sim p_{\mathrm{data}}(x\mid c)\) and \(Z_t=tX+(1-t)\epsilon\), where \(\epsilon\sim\mathcal{N}(0,I)\) is independent of \(X\), and \(t=0\) denotes noise while \(t=1\) denotes clean data. JiT directly predicts the clean image by minimizing
\begin{equation}
    \mathcal{L}_{\mathrm{JiT}}
    =
    \mathbb{E}\!\left[
        w(t)\,
        \|f_\theta(Z_t,t,c)-X\|_2^2
    \right],
    \qquad
    w(t)=\frac{1}{(1-t)^2},
    \label{eq:jit_objective}
\end{equation}
where the expectation is over the data, conditioning, corruption noise, and the JiT time distribution \(t\sim p(t)\) on \((0,1)\). Writing \(U_t=(Z_t,t,c)\) for the student input, the population regression target is \(m_t(Z_t,c)=\mathbb{E}[X\mid U_t]\).

\subsection{Auxiliary Supervision in JiT}
\label{sec:auxiliary_jit}
Conventional feature alignment supervises an intermediate representation \(h_\theta^\ell(Z_t,t,c)\) toward a feature target \(\phi(X)\). Under squared regression, the corresponding conditional statistic is
\begin{equation}
    r_t(Z_t,c)
    =
    \mathbb{E}[\phi(X)\mid U_t],
    \qquad
    r_t\not\equiv \phi\circ m_t
    \quad\text{in general}.
    \label{eq:feature_mismatch}
\end{equation}
A sufficiently expressive representation may support both statistics. Their interaction nevertheless merits examination in JiT: DiT-B/2 maps a \(16\)-dimensional latent patch into a \(768\)-dimensional hidden token, whereas JiT-B/16 maps \(768\) pixel values into the same width. These dimensions do not determine effective model capacity, but motivate studying how auxiliary feature learning interacts with the representational demands of clean-image prediction.

\paragraph{Training behavior.}
\begin{wrapfigure}{r}{0.4\textwidth}
    \vspace{-2.5em}
    \centering
    \includegraphics[width=\linewidth]{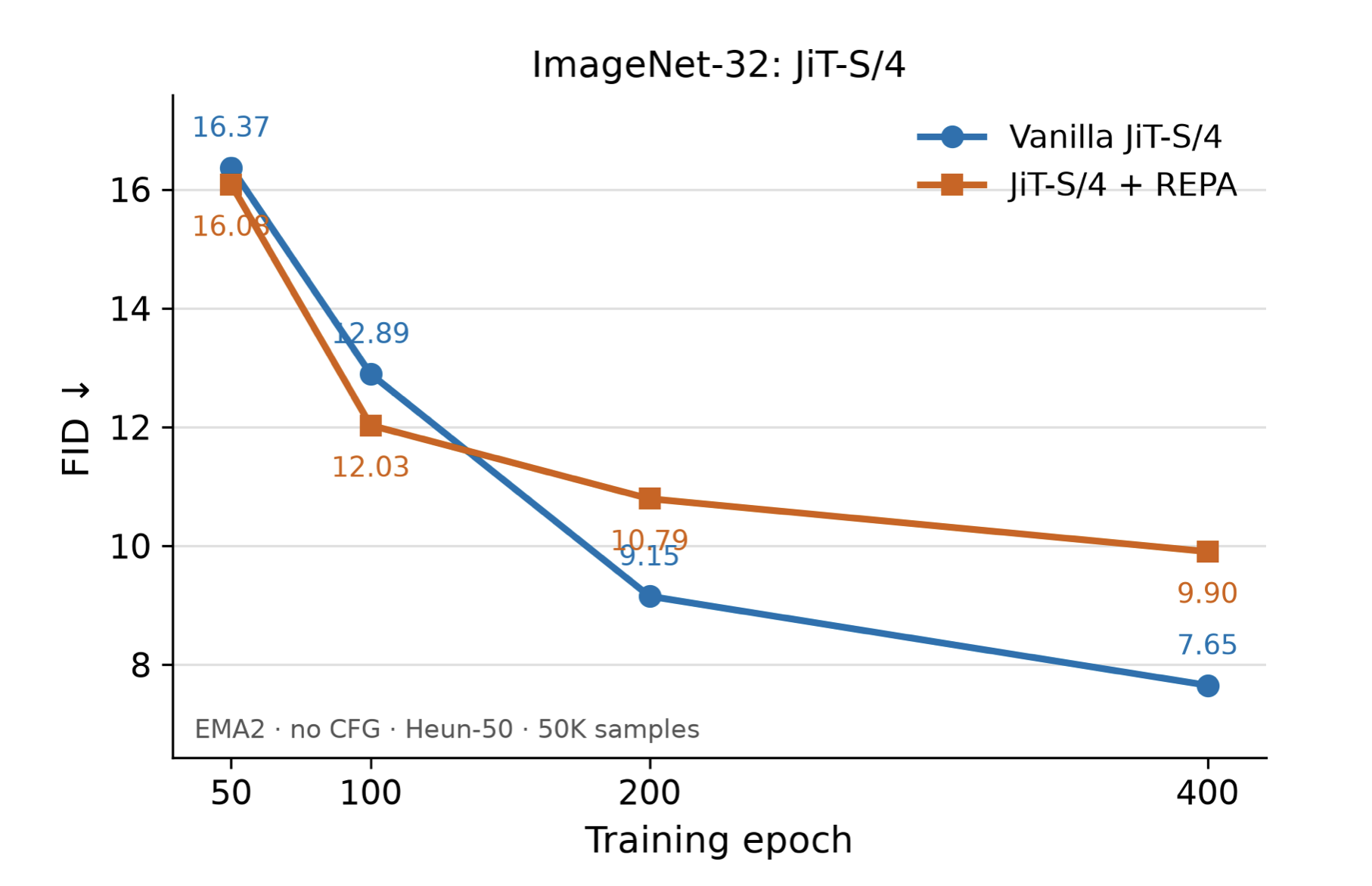}\\
    \vspace{-0.15in}
    \caption{REPA initially improves FID on JiT-S/4 at ImageNet \(32\times32\), but the advantage reverses later in training.}
    \label{fig:imagenet32}
    \vspace{-1.0em}
\end{wrapfigure}
PixelREPA~\citep{shin2026representation} relates the difficulty of applying REPA in pixel space to the mismatch between high-resolution pixels and compressed semantic features, reporting improvements at ImageNet \(32\times32\) but later degradation at \(256\times256\). We revisit the low-resolution regime with a smaller JiT-S/4 (Fig.~\ref{fig:imagenet32}). REPA improves FID from \(16.37\) to \(16.08\) at 50 epochs and from \(12.89\) to \(12.03\) at 100 epochs. At 200 and 400 epochs, however, it obtains FIDs of \(10.79\) and \(9.90\), compared with \(9.15\) and \(7.65\) for vanilla JiT. Thus, reducing resolution does not by itself ensure sustained gains from feature alignment. Our JiT-B/16 comparisons also show limited gains from directly transferred hidden self-alignment and regularization methods (Table~\ref{tab:alignment_comparison}). These observations motivate examining the interaction between auxiliary feature learning and clean-image prediction in the tested JiT settings.

\begin{figure}[t]
    \centering
    \includegraphics[width=\linewidth]{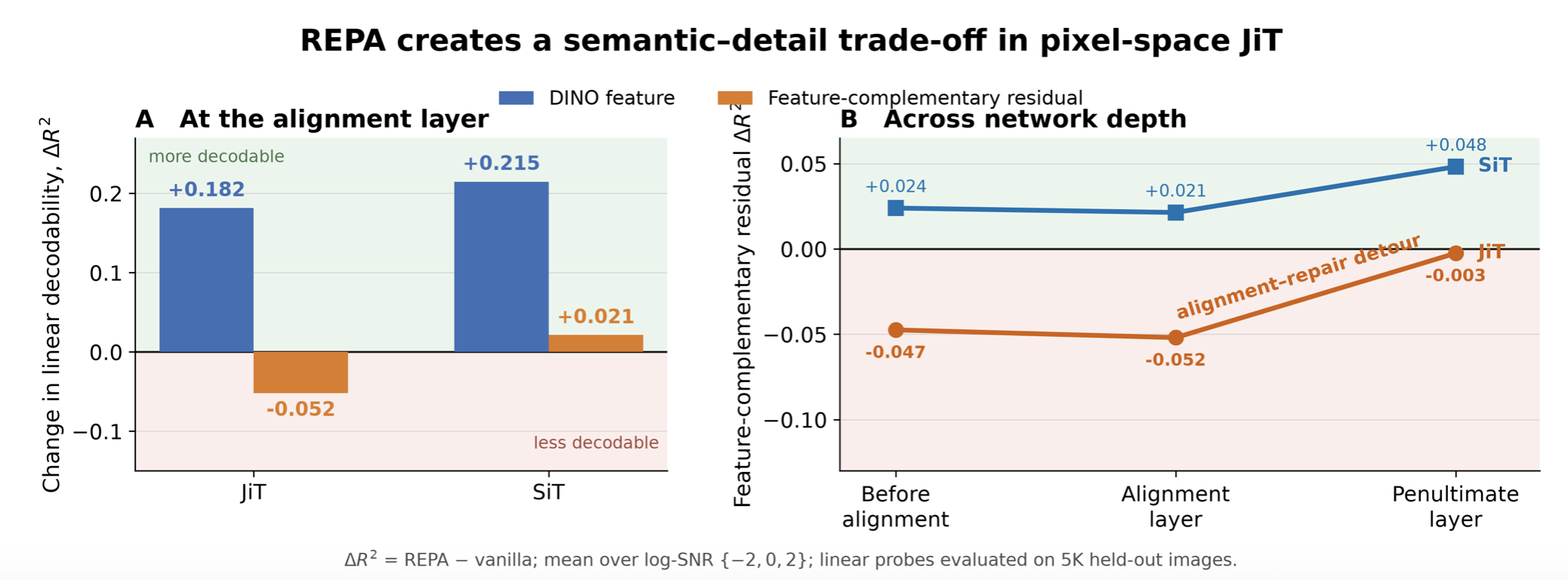}\\
    \vspace{-0.1in}
    \caption{Linear-probe observations at epoch 200. (A) REPA increases DINO-feature decodability in both JiT and SiT, while feature-complementary residual decodability decreases in JiT and increases in SiT. (B) JiT exhibits an \emph{alignment--repair detour} in linear decodability: its residual deficit narrows toward the output. SiT shows gains at all measured layers.}
    \vspace{-0.15in}
    \label{fig:information_allocation}
\end{figure}

\paragraph{Linear-probe observations.}
We probe intermediate representations at epoch 200 using two clean-image targets: the DINO feature \(\phi(X)\)~\citep{oquab2024dinov2}, and the feature-complementary residual \(X_\perp=X-\Pi_\phi X\), where \(\Pi_\phi X\) is the linear prediction of \(X\) from \(\phi(X)\). The latter measures access to clean-image variation not linearly explained by the aligned features. Figure~\ref{fig:information_allocation} reports changes in probe \(R^2\) relative to vanilla models, averaged over log-SNR levels \(\{-2,0,2\}\).
At the alignment layer, REPA increases DINO-feature decodability by \(0.182\) in JiT and \(0.215\) in SiT~\citep{ma2024sit}. Residual decodability, however, decreases by \(0.052\) in JiT and increases by \(0.021\) in SiT. Across depth, JiT's residual decodability changes are \(-0.047\) before the alignment layer, \(-0.052\) at that layer, and \(-0.003\) at the penultimate layer; SiT shows positive changes at all three layers.

We refer to this pattern as an \emph{alignment--repair detour}: the deficit in linear decodability of feature-complementary variation at intermediate layers largely diminishes toward the output. This pattern concerns linear accessibility and does not by itself establish information loss or an additional optimization cost. These observations are consistent with a capacity-sensitive trade-off in the evaluated JiT setting: greater linear accessibility of the aligned features accompanies reduced accessibility of complementary clean-image variation. They provide an empirical motivation to study auxiliary supervision whose relation to the denoising target can be analyzed directly.

\subsection{What Should an Auxiliary Prediction Target Preserve?}
\label{sec:target_requirements}
These observations motivate auxiliary supervision directly on the clean-image prediction. To keep this supervision aligned with JiT's denoising task, a target \(Y\) with finite second moments should satisfy \emph{conditional target consistency}:
\begin{equation}
    \mathbb{E}[Y\mid U_t]
    =
    \mathbb{E}[X\mid U_t]
    =
    m_t(Z_t,c).
    \label{eq:compatible_target}
\end{equation}
This condition preserves the population regression target required by JiT.

Consistency alone, however, does not explain how an auxiliary target can improve training: the clean-image target \(X\) already satisfies this condition. A potential benefit lies in reducing the variability of the supervision. Conditional on a noisy input \(U_t\), clean-image targets vary around the posterior mean; under squared regression, this target variation contributes to stochastic-gradient variability. We therefore seek a target that retains the same conditional mean while reducing conditional target variance, expressed by the covariance ordering
\begin{equation}
    \operatorname{Cov}(Y\mid U_t)
    \preceq
    \operatorname{Cov}(X\mid U_t),
    \label{eq:target_covariance}
\end{equation}
where \(\preceq\) denotes the positive-semidefinite order. This provides a possible optimization benefit while preserving the underlying regression task.

In Section~\ref{sec:method}, we construct a target from a cleaner, Markov-coupled observation and establish both properties for oracle clean-image predictions. We then address the approximation error of an actual EMA teacher through ground-truth supervision and reliability-gated pair selection.

\section{Method}
\label{sec:method}

We present~\ourwork, which supervises clean-image predictions across Markov-coupled noise levels. We first construct an oracle target satisfying the conditional consistency and variance criteria of Section~\ref{sec:target_requirements}, then instantiate the method with an EMA teacher and reliability-gated pair selection.

\subsection{Markov-Coupled Observations}
\label{sec:msa}
\begin{figure}[t]
    \centering
    \includegraphics[width=\linewidth]{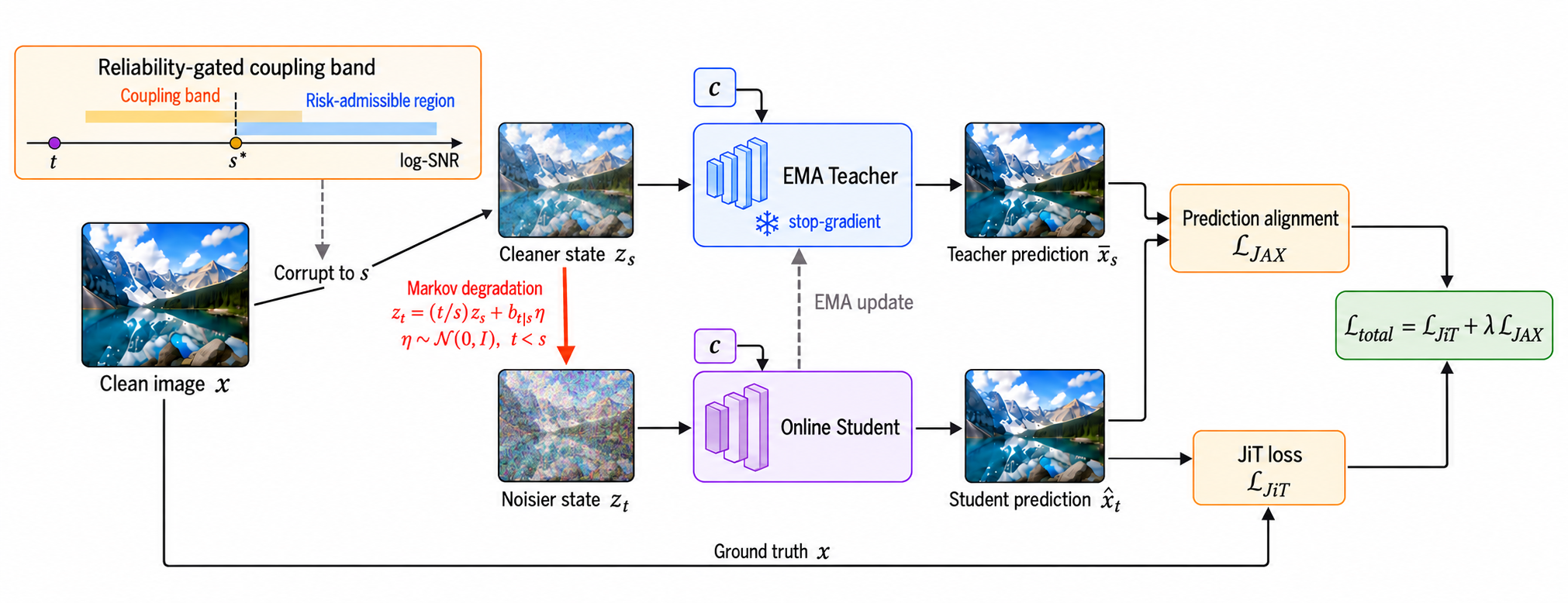}\\
    \vspace{-0.15in}
    \caption{Overview of~\ourwork. An EMA teacher predicts the clean image from a cleaner observation, while the online student receives a Markov-coupled noisier observation. Training combines direct JiT supervision with prediction alignment, without modifying the architecture or sampling procedure.}
    \vspace{-0.1in}
    \label{fig:method_overview}
\end{figure}

We construct the cleaner and noisier observations through a Markov degradation that preserves the student's original corruption marginal. Figure~\ref{fig:method_overview} illustrates the resulting training pipeline.

Let $0<t<s<1$, so that $Z_s$ is the cleaner state. We first corrupt the clean image and then apply an incremental Gaussian degradation:
\begin{equation}
    Z_s=sX+(1-s)\epsilon_s,\quad
    Z_t=a_{t|s}Z_s+b_{t|s}\eta,\quad
    a_{t|s}=\tfrac{t}{s},\quad
    b_{t|s}^2=(1-t)^2-\tfrac{t^2}{s^2}(1-s)^2.
    \label{eq:coupling}
\end{equation}
Here $\epsilon_s,\eta\sim\mathcal{N}(0,I)$ are independent of each other and of $(X,c)$, and both predictions use the same conditioning $c$. This transition is well defined for $t<s$ and preserves the JiT marginal, $Z_t\mid(X=x,c)\sim\mathcal{N}(tx,(1-t)^2I)$, while inducing the Markov chain
\begin{equation}
    X\;\longrightarrow\;Z_s\;\longrightarrow\;Z_t
    \qquad\text{conditional on }(c,t,s).
    \label{eq:markov}
\end{equation}
Thus, the student receives the same marginally corrupted input as in standard JiT, now coupled to a cleaner observation. This Markov structure yields the conditional consistency established below; independent corruptions of the same image do not generally provide this property. Appendix~\ref{app:markov-martingale} verifies the construction.

\subsection{Oracle Properties}
\label{sec:oracle}
Recall that $U_t=(Z_t,t,c)$ and $m_t(Z_t,c)=\mathbb{E}[X\mid U_t]$. Under the Markov coupling, the cleaner posterior prediction provides a conditionally consistent target for the student.

\begin{proposition}[Martingale consistency]
\label{prop:mart}
For any $0<t<s<1$, the oracle clean-image predictions satisfy
\begin{equation}
    \mathbb{E}\bigl[m_s(Z_s,c)\mid U_t,s\bigr]=m_t(Z_t,c).
    \label{eq:mart}
\end{equation}
\end{proposition}
The cleaner prediction can exploit information unavailable in $Z_t$, so the two predictions need not agree for each paired sample. Averaging over cleaner observations compatible with the student input nevertheless recovers exactly its posterior mean. This conditional consistency gives the posterior predictions their martingale structure, formalized in Appendix~\ref{app:markov-martingale}.

Let $\pi(t,s)=p(t)q(s\mid t)$ be a pair distribution supported on $t<s$, with the JiT student-time marginal $p(t)$. The pair is sampled independently of $(X,c)$ and the corruption noises. Using the same weight $w(t)$ as JiT, define the oracle objective
\begin{equation}
    \mathcal{L}^{\star}_{\mathrm{JA}x}(\theta)
    =\mathbb{E}_{\pi}\!\left[
        w(t)\bigl\|f_\theta(U_t)-\mathrm{sg}[m_s(Z_s,c)]\bigr\|_2^2
    \right],
    \label{eq:oracle}
\end{equation}
where $\mathrm{sg}$ denotes stop-gradient and the expectation also includes the data and coupled corruptions. Proposition~\ref{prop:mart} implies
\begin{equation}
    \mathcal{L}^{\star}_{\mathrm{JA}x}(\theta)
    =\mathcal{L}_{\mathrm{JiT}}(\theta)+K,
    \qquad
    \nabla_\theta\mathcal{L}^{\star}_{\mathrm{JA}x}
    =\nabla_\theta\mathcal{L}_{\mathrm{JiT}},
    \label{eq:equiv}
\end{equation}
where $K$ is independent of $\theta$. The oracle objective therefore preserves the population loss up to a constant, including its minimizers and expected gradient.

Although the population gradients agree, the regression targets have different conditional variability. The Markov coupling yields the following covariance ordering.
\begin{proposition}[Variance reduction]
\label{prop:var}
For any fixed $0<t<s<1$,
\begin{equation}
    \operatorname{Cov}\bigl(m_s(Z_s,c)\mid U_t\bigr)
    =\operatorname{Cov}(X\mid U_t)
    -\mathbb{E}\bigl[\operatorname{Cov}(X\mid Z_s,c)\mid U_t\bigr]
    \preceq \operatorname{Cov}(X\mid U_t).
    \label{eq:var}
\end{equation}
For fixed $t$, the covariance reduction is nonincreasing as $s$ increases. The oracle target approaches $m_t(Z_t,c)$ as $s\downarrow t$ and $X$ as $s\uparrow1$, with both limits in mean square.
\end{proposition}

The oracle target removes the posterior uncertainty that remains unresolved at the cleaner observation while retaining the conditional mean required by JiT. Thus, it satisfies both criteria in Section~\ref{sec:target_requirements}, and $K\le0$. For fixed student parameters, replacing $X$ with this oracle target also preserves the expected gradient and does not increase its covariance. Proofs, endpoint limits, and the corresponding gradient analysis are given in Appendices~\ref{app:objective-variance} and~\ref{app:stochastic-gradients}.

\subsection{Practical Prediction Supervision}
\label{sec:practical}
The oracle predictor $m_s$ is unavailable in practice. We approximate it with an exponential-moving-average (EMA) teacher, whose parameters $\bar\theta$ track the online student. Given the cleaner observation, the teacher produces the target $\bar{x}_s=\mathrm{sg}[f_{\bar\theta}(Z_s,s,c)]$.

Let $a_k(t)\in\{0,1\}$ indicate whether alignment is active at update $k$, and let $s^\star_k(t)$ be the selected teacher time, as specified in Section~\ref{sec:band}. With the teacher and selection rule held fixed during the student update, we combine prediction alignment with direct clean-image supervision:
\begin{equation}
    \mathcal{L}_{\mathrm{JA}x,k}(\theta)
    =\mathbb{E}\!\left[
        a_k(t)w(t)\bigl\|f_\theta(U_t)-\bar{x}_{s^\star_k(t)}\bigr\|_2^2
    \right],\quad
    \mathcal{L}_{\mathrm{total},k}
    =\mathcal{L}_{\mathrm{JiT}}+\lambda\mathcal{L}_{\mathrm{JA}x,k}.
    \label{eq:total}
\end{equation}
The alignment integrand is zero for inactive times, without requiring a teacher prediction.

Approximation error can shift the teacher target's conditional mean and alter its covariance, so the EMA objective does not automatically inherit the oracle guarantees. Direct clean-image supervision attenuates the teacher-induced shift in the conditional regression target. We further restrict alignment to empirically reliable teacher states through the selection rule below. Appendix~\ref{app:ema-reliability} analyzes the teacher bias and the effective target and weighting of the combined objective.

\subsection{Reliability-Gated Pair Selection}
\label{sec:band}
The oracle analysis favors nearby teacher states, which provide greater variance reduction. As $s$ approaches $t$, however, the conditional bias of an EMA teacher need not vanish: alignment approaches same-state EMA self-distillation, whose usefulness depends on teacher accuracy. At the opposite endpoint, the oracle target approaches $X$ as $s\to1$, leaving little variance reduction. We therefore seek an empirically reliable teacher within a bounded separation from the student.

\paragraph{Coupling band.}
We parameterize time by the log-SNR $\ell_t=2\log\frac{t}{1-t}$. Under Eq.~\ref{eq:coupling}, the correlation between the standardized corruption noises is
\begin{equation}
    r(t,s)=\frac{t(1-s)}{s(1-t)}
    =\exp\!\left(-\frac{\ell_s-\ell_t}{2}\right).
    \label{eq:corr}
\end{equation}
The gap $\Delta\ell=\ell_s-\ell_t$ therefore controls the shared corruption noise. We constrain active pairs to $\Delta_{\min}\le\Delta\ell\le\Delta_{\max}$, with $\Delta_{\min}>0$, to avoid nearly identical or excessively separated states.

\paragraph{Empirical reliability frontier.}
We periodically estimate the teacher's mean squared clean-image prediction risk $R_k(\ell)$ by log-SNR. Using estimates available before the current batch, we define
\begin{equation}
    \ell_{\mathrm{safe}}(k)
    =\inf\{\ell:R_k(\ell)\le R^\star\}+\delta_{\mathrm{safe}},
    \label{eq:frontier}
\end{equation}
where $R^\star$ is the admissible risk and $\delta_{\mathrm{safe}}$ is a margin toward cleaner states. This frontier approximates an admissible region on the cleaner side of the risk crossing. Prediction risk motivates this empirical reliability criterion through its relation to teacher bias (Appendix~\ref{app:ema-reliability}).

For a student time $t$, we choose
\begin{equation}
    \ell_s^\star=\max\{\ell_t+\Delta_{\min},\,\ell_{\mathrm{safe}}(k)\},
    \qquad
    s^\star_k(t)=\operatorname{sigmoid}(\ell_s^\star/2).
    \label{eq:teacher_selection}
\end{equation}
We set $a_k(t)=1$ only if the frontier is available, $\ell_s^\star-\ell_t\le\Delta_{\max}$, and $\ell_s^\star\le\ell_{\max}$; otherwise, the sample receives only JiT supervision. Within this frontier-defined feasible set, the selected state is the nearest admissible teacher and maximizes the oracle covariance reduction. This motivates the rule as an empirical response to the bias--variance trade-off; its effectiveness with an EMA teacher is evaluated experimentally.

\paragraph{Batch objective.}
Let $\mathcal{A}_k=\{i:a_k(t_i)=1\}$ denote the active samples in a batch of size $B$. The alignment loss is
\begin{equation}
    \widehat{\mathcal{L}}_{\mathrm{JA}x,k}
    =\frac{1}{B}\sum_{i\in\mathcal{A}_k}
    \frac{\bigl\|f_\theta(Z_{t_i},t_i,c_i)-\bar{x}_{s^\star_k(t_i)}\bigr\|_2^2}
    {(1-t_i)^2}.
    \label{eq:batched}
\end{equation}
Normalization uses the full batch size, so inactive samples contribute zero alignment loss and each active sample retains its weight as the activation fraction changes. Training pseudocode (Algorithm~\ref{alg:jax}), calibration settings, and other implementation details are provided in Appendix~\ref{app:implementation}.

\section{Experiments}
\label{sec:experiments}

\begin{wrapfigure}{r}{0.58\textwidth}
    \vspace{-2em}
    \centering
\includegraphics[width=\linewidth]{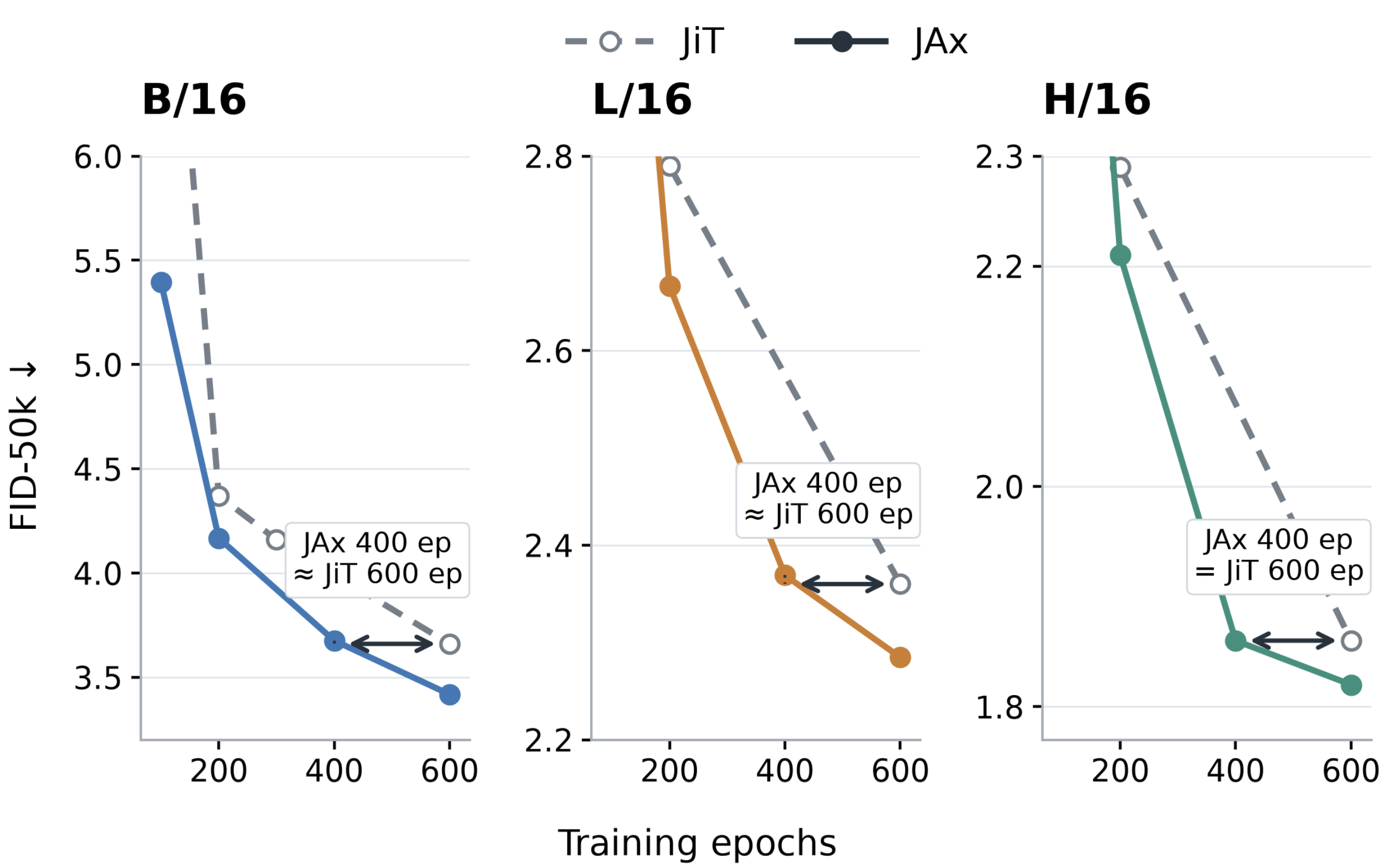}\\
    \vspace{-0.15in}
    \caption{Training convergence across JiT scales. Dashed and solid lines denote JiT and \ourwork, respectively. JiT-L/16 and H/16 baselines use the reported 200- and 600-epoch results.}
\label{fig:training_acceleration}
\vspace{-1em}
\end{wrapfigure}
\paragraph{Experimental setup.} We evaluate on ImageNet $256\times256$ using the JiT-B/16, L/16, and H/16 architectures and base training recipe~\citep{li2026back}, with a shared alignment weight $\lambda=0.5$. We report FID~\citep{heusel2017gans} and Inception Score (IS)~\citep{salimans2016improved} using 50K generated samples. Sampling follows JiT with a 50-step Heun ODE solver and CFG~\citep{ho2022classifier} over the interval $[0.1,1]$~\citep{kynkaanniemi2024applying}. Full training, EMA, and evaluation settings are provided in Appendix~\ref{app:implementation}.

\paragraph{Generation quality and convergence.} Figure~\ref{fig:training_acceleration} shows improved FID and faster convergence in training epochs across model scales. At 600 epochs, \ourwork reduces FID from 3.66 to 3.41 on B/16, from 2.36 to 2.26 on L/16, and from 1.86 to 1.82 on H/16. It also reaches approximately the corresponding JiT model's 600-epoch FID at 400 epochs. These results support the effectiveness of prediction supervision across the evaluated scales. Full metrics and training-cost analysis are provided in Appendix~\ref{app:metrics-cost}; broader system-level comparisons appear in Appendix~\ref{app:system-comparison}.

\begin{wraptable}{r}{0.52\textwidth}
    \vspace{-1em}
    \centering
    \small
    \setlength{\tabcolsep}{4pt}
    \renewcommand{\arraystretch}{1.12}
    \caption{Auxiliary supervision on JiT-B/16 after 200 epochs. External representations are supplied by a separate pretrained encoder. Bold and underlining mark the best FID overall and without external representations, respectively.}
    \label{tab:alignment_comparison}
    \begin{tabular*}{\linewidth}{@{\extracolsep{\fill}}|l|c|r|}
        \hline
        \rowcolor{gray!15}
        \textbf{Method} & \shortstack{\textbf{External representations}} & \textbf{FID $\downarrow$} \\
        \hline
        JiT & None & 4.37 \\
        \hline
        JiT + SRA 2 & VAE & 4.48 \\
        JiT + SRA & None & 4.38 \\
        JiT + Self-Flow & None & 4.31 \\
        JiT + REPA & DINOv2 & 5.12 \\
        \hline
        \rowcolor{gray!8}
        \ourwork (ours) & None & \underline{4.17} \\
        PixelREPA & DINOv2 & 4.00 \\
        \rowcolor{gray!8}
        PixelREPA + \ourwork & DINOv2 & \textbf{3.91} \\
        \hline
    \end{tabular*}
    \vspace{-1em}
\end{wraptable}

\paragraph{Comparison with feature alignment methods.} Table~\ref{tab:alignment_comparison} compares auxiliary supervision methods on JiT-B/16 after 200 epochs. Directly transferring existing methods yields limited gains: SRA 2, SRA, and REPA underperform vanilla JiT, while Self-Flow improves FID from 4.37 to 4.31. By supervising clean-image predictions, \ourwork achieves 4.17 without an external encoder, supporting our motivation to align auxiliary supervision with the denoising target. PixelREPA achieves a lower FID of 4.00 using external semantic supervision. Combining it with \ourwork further improves FID to 3.91, supporting the compatibility of prediction supervision and pixel-specific feature alignment in this setting.

\begin{wrapfigure}{r}{0.54\textwidth}
    \vspace{-2em}
    \centering
    \includegraphics[width=\linewidth]{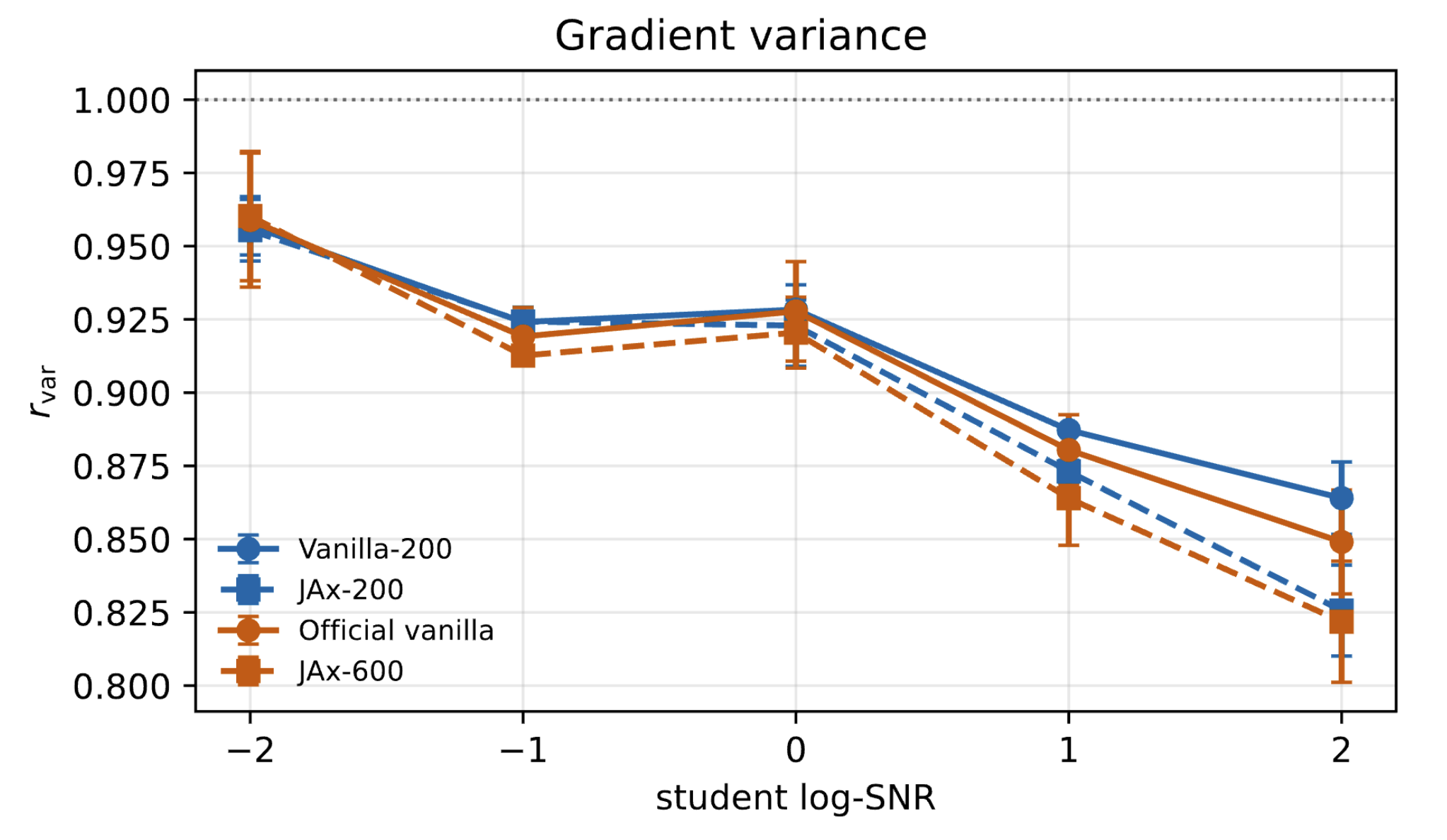}\\
    \vspace{-0.15in}
    \caption{Minibatch gradient variance relative to JiT at epoch 200 and epoch 600, after normalizing the combined \ourwork gradient by $1+\lambda$. \ourwork has lower gradient variance across training stages.}
    \vspace{-0.1in}
    \label{fig:gradient_variance}
\end{wrapfigure}
\paragraph{Optimization behavior.} 
Figure~\ref{fig:gradient_variance} compares paired minibatch gradients under JiT supervision and the combined \ourwork objective, normalized by $1+\lambda$ to remove the overall scale difference. We evaluate our epoch-200 vanilla and \ourwork checkpoints, together with an official vanilla JiT checkpoint and our 600-epoch \ourwork checkpoint, using their corresponding epoch EMA teacher. All paired samples used in this diagnostic have active prediction alignment,
i.e., $a_k(t)=1$. Normalized \ourwork gradients exhibit lower variance at every evaluated noise level, with an average reduction of $9.6\%$ across noise levels and student states ($9.4\%$ at epoch 200 and $9.9\%$ in the 600-epoch comparison). Accounting also for the mean-gradient shift, the estimated mean-squared deviation from the corresponding JiT population gradient decreases by $6.5\%$ ($6.4\%$ at epoch 200 and $6.6\%$ for 600). These measurements support the variance-reduction motivation of our oracle analysis in the tested EMA setting; the decomposition and theoretical scope are given in Appendix~\ref{app:stochastic-gradients}.

\begin{figure}[t]
    \centering
    \includegraphics[width=\textwidth]{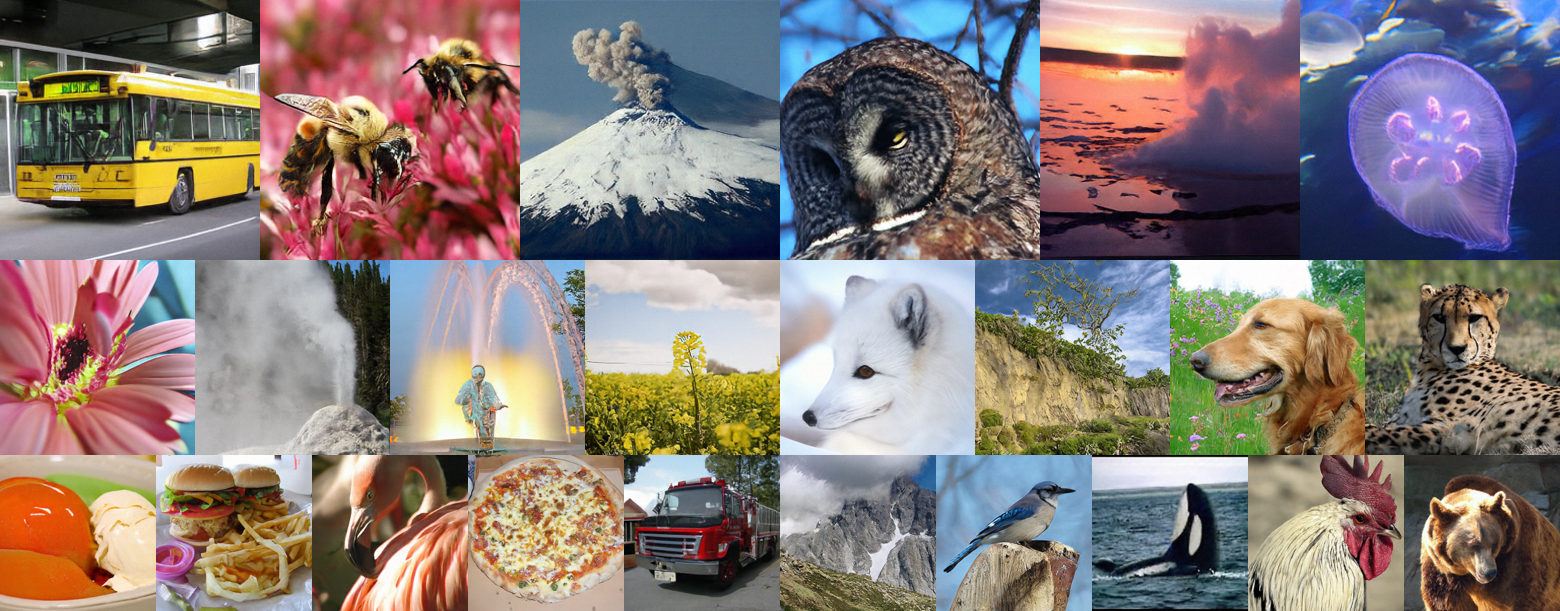}\\
    \vspace{-0.1in}
    \caption{Curated samples generated by \ourwork on ImageNet $256\times256$ using the standard JiT-H architecture and sampling procedure.}
    \vspace{-0.15in}
    \label{fig:qualitative}
\end{figure}

\paragraph{Qualitative results.}
Figure~\ref{fig:qualitative} presents samples generated by \ourwork across diverse ImageNet categories. These examples illustrate the visual quality of the generated images and complement the quantitative improvements in FID. The reported FID gains are achieved by modifying only the prediction supervision during training, while retaining the JiT-H architecture and the original sampling procedure, with no additional inference cost.

\section{Ablation Study}
\label{sec:ablation}

We examine teacher-state selection, observation coupling, and the distinction between target modification and time reweighting on JiT-B/16. Table~\ref{tab:pair_coupling_ablation} summarizes 200-epoch results on ImageNet $256\times256$.

\paragraph{Reliability-gated pair selection.} Fixed-gap alignment at $\Delta\ell=2$, enabled after warmup, and conditional sampling from the JiT time distribution restricted to $s>t$ yield FIDs of 4.48 and 4.43, respectively, compared with 4.37 for vanilla JiT. The fixed-gap control retains the same EMA, alignment weight, time-feasibility constraints, and full-batch normalization as \ourwork. Every evaluated reliability-gated band improves upon the vanilla baseline. With the minimum gap fixed at 2, tightening the upper bound from 4 to 2.5 improves FID from 4.29 to 4.17. These results support selecting nearby teacher states through an empirical reliability criterion: neither Markov coupling alone nor fixed-gap alignment reproduces the gains of reliability-gated selection in this setting.

\paragraph{Markov coupling.} Within the same reliability-gated band $[2,2.5]$, Markov coupling achieves 4.17 FID, compared with 4.49 for the non-Markov variant, using the same pair-selection rule, teacher configuration, and alignment weight. This comparison supports our choice of Markov-coupled prediction supervision and shows that reliability-gated selection alone is insufficient in the tested setting.
\begin{wraptable}{r}{0.52\textwidth}
    \vspace{-1.5em}
    \centering
    \small
    \setlength{\tabcolsep}{4pt}
    \renewcommand{\arraystretch}{1.10}
    \caption{Ablations on JiT-B/16 at 200 epochs. RCB denotes the reliability-gated coupling band. Prediction-supervision variants use Markov coupling unless noted. FID uses 50K samples.}
    \label{tab:pair_coupling_ablation}
    \begin{tabular*}{\linewidth}{@{\extracolsep{\fill}}|lcc|}
        \hline
        \rowcolor{gray!15}
        Method & $\Delta\ell$ & FID $\downarrow$ \\
        \hline
        \rowcolor{gray!8}
        \multicolumn{3}{|l|}{\textbf{Baseline and controls}} \\
        \hline
        Vanilla JiT & -- & 4.37 \\
        Reweighting only & -- & 4.39 \\
        Fixed gap & $2$ & 4.48 \\
        Conditional sampling & $>0$ & 4.43 \\
        \hline
        \rowcolor{gray!8}
        \multicolumn{3}{|l|}{\textbf{Reliability-gated bands}} \\
        \hline
        RCB & $[2,4]$ & 4.29 \\
        RCB & $[1.5,4]$ & 4.32 \\
        RCB & $[2,3.5]$ & 4.26 \\
        RCB & $[2,3]$ & 4.23 \\
        RCB (\ourwork) & $[2,2.5]$ & \textbf{4.17} \\
        \hline
        \rowcolor{gray!8}
        \multicolumn{3}{|l|}{\textbf{Observation coupling}} \\
        \hline
        RCB (Non-Markov) & $[2,2.5]$ & 4.49 \\
        \hline
    \end{tabular*}
    \vspace{-4em}
\end{wraptable}
\paragraph{Teacher targets and time reweighting.} The combined objective changes both the regression target and the effective time weighting (Appendix~\ref{app:ema-reliability}). To isolate the teacher target's contribution, the reweighting-only control replays the saved \ourwork activation schedule $a_k(t)$ and optimizes $w(t)(1+\lambda a_k(t))\|f_\theta(U_t)-X\|_2^2$, with initialization, data order, and the base training protocol matched. It yields 4.39 FID, compared with 4.37 for vanilla JiT and 4.17 for \ourwork. Thus, time reweighting alone does not reproduce the improvement, supporting the contribution of teacher targets beyond the induced weighting in this setting.

\paragraph{Summary of the appendix.} Appendix A provides proofs and further analysis of Markov coupling, variance reduction, and teacher approximation error. Appendices B–D discuss related methods, implementation details, and limitations. Appendices E–F provide additional quantitative comparisons, cost analysis, and qualitative results.

\section{Conclusion}
\label{sec:conclusion}

We presented \ourwork, a prediction-supervision method for JiT motivated by the observed limitations of hidden representation alignment. Under Markov coupling, oracle clean-image predictions preserve the required conditional mean and reduce conditional target covariance, with a population loss equivalent to JiT's up to a constant. The practical method combines an EMA teacher with direct clean-image supervision and reliability-gated pair selection to address teacher approximation error. Experiments demonstrate improved final FID and faster convergence in training epochs across JiT scales, without external encoders or changes to the architecture and sampling procedure. Ablations show that time reweighting alone and fixed-gap alignment after warmup do not reproduce these gains, supporting the contributions of both teacher targets and reliability-gated selection. Gradient diagnostics further show lower normalized gradient variance and mean-squared deviation from the JiT population gradient. Combining \ourwork with PixelREPA also improves FID in the evaluated JiT-B/16 setting.




\newpage
\bibliography{iclr2025_conference}

@inproceedings{daras2023consistent,
  title={Consistent Diffusion Models: Mitigating Sampling Drift by Learning to be Consistent},
  author={Daras, Giannis and Dagan, Yuval and Dimakis, Alexandros G. and Daskalakis, Constantinos},
  booktitle={Advances in neural information processing systems (NeurIPS)},
  year={2023},
}

@inproceedings{song2023consistency,
  title={Consistency Models},
  author={Song, Yang and Dhariwal, Prafulla and Chen, Mark and Sutskever, Ilya},
  booktitle={International Conference on Machine Learning (ICML)},
  year={2023},

}

@article{yang2026stable,
  title={Stable Velocity: A Variance Perspective on Flow Matching},
  author={Yang, Donglin and Zhang, Yongxing and Yu, Xin and Hou, Liang and Tao, Xin and Wan, Pengfei and Qi, Xiaojuan and Liao, Renjie},
  journal={arXiv preprint arXiv:2602.05435},
  year={2026},

}

@inproceedings{peebles2023scalable,
  title={Scalable diffusion models with transformers},
  author={Peebles, William and Xie, Saining},
  booktitle={IEEE/CVF International Conference on Computer Vision (ICCV)},
  pages={4172--4182},
  year={2023},
}

@inproceedings{ma2024sit,
  title={{SiT}: Exploring flow and diffusion-based generative models with scalable interpolant transformers},
  author={Ma, Nanye and Goldstein, Mark and Albergo, Michael S. and Boffi, Nicholas M. and Vanden-Eijnden, Eric and Xie, Saining},
  booktitle={European Conference on Computer Vision (ECCV)},
  pages={23--40},
  year={2024},
  organization={Springer}
}

@inproceedings{rombach2022high,
  title={High-resolution image synthesis with latent diffusion models},
  author={Rombach, Robin and Blattmann, Andreas and Lorenz, Dominik and Esser, Patrick and Ommer, Bj{\"o}rn},
  booktitle={Proceedings of the IEEE/CVF conference on computer vision and pattern recognition},
  pages={10684--10695},
  year={2022}
}

@inproceedings{song2020score,
  title={Score-based generative modeling through stochastic differential equations},
  author={Song, Yang and Sohl-Dickstein, Jascha and Kingma, Diederik P and Kumar, Abhishek and Ermon, Stefano and Poole, Ben},
  booktitle={International Conference on Learning Representations (ICLR)},
  year={2021}
}

@inproceedings{ho2020denoising,
  title={Denoising diffusion probabilistic models},
  author={Ho, Jonathan and Jain, Ajay and Abbeel, Pieter},
  booktitle={Advances in neural information processing systems (NeurIPS)},
  pages={6840--6851},
  year={2020}
}

@inproceedings{lipman2022flow,
  title={Flow matching for generative modeling},
  author={Lipman, Yaron and Chen, Ricky TQ and Ben-Hamu, Heli and Nickel, Maximilian and Le, Matt},
  booktitle={International Conference on Learning Representations (ICLR)},
  year={2023}
}

@inproceedings{liu2022flow,
  title={Flow straight and fast: Learning to generate and transfer data with rectified flow},
  author={Liu, Xingchao and Gong, Chengyue and Liu, Qiang},
  booktitle={International Conference on Learning Representations (ICLR)},
  year={2023}
}

@inproceedings{yu2024representation,
  title={Representation alignment for generation: Training diffusion transformers is easier than you think},
  author={Yu, Sihyun and Kwak, Sangkyung and Jang, Huiwon and Jeong, Jongheon and Huang, Jonathan and Shin, Jinwoo and Xie, Saining},
  booktitle={International Conference on Learning Representations (ICLR)},
  year={2025}
}

@inproceedings{singh2025matters,
  title={What matters for representation alignment: Global information or spatial structure?},
  author={Singh, Jaskirat and Leng, Xingjian and Wu, Zongze and Zheng, Liang and Zhang, Richard and Shechtman, Eli and Xie, Saining},
  booktitle={International Conference on Learning Representations (ICLR)},
  year={2026}
}

@article{wang2026sra,
  title={{SRA} 2: Variational autoencoder self-representation alignment for efficient diffusion training},
  author={Wang, Mengmeng and Jiang, Dengyang and Li, Liuzhuozheng and Lin, Yucheng and Shen, Guojiang and Kong, Xiangjie and Liu, Yong and Dai, Guang and Wang, Jingdong},
  journal={arXiv preprint arXiv:2601.17830},
  year={2026}
}

@inproceedings{jiang2025no,
  title={No other representation component is needed: Diffusion transformers can provide representation guidance by themselves},
  author={Jiang, Dengyang and Wang, Mengmeng and Li, Liuzhuozheng and Zhang, Lei and Wang, Haoyu and Wei, Wei and Dai, Guang and Zhang, Yanning and Wang, Jingdong},
  booktitle={International Conference on Learning Representations (ICLR)},
  year={2026}
}

@article{chefer2026self,
  title={Self-supervised flow matching for scalable multi-modal synthesis},
  author={Chefer, Hila and Esser, Patrick and Lorenz, Dominik and Podell, Dustin and Raja, Vikash and Tong, Vinh and Torralba, Antonio and Rombach, Robin},
  journal={arXiv preprint arXiv:2603.06507},
  year={2026}
}

@article{wang2025diffuse,
  title={Diffuse and disperse: Image generation with representation regularization},
  author={Wang, Runqian and He, Kaiming},
  journal={arXiv preprint arXiv:2506.09027},
  year={2025}
}

@article{singh2026improved,
  title={Improved baselines with representation autoencoders},
  author={Singh, Jaskirat and Zheng, Boyang and Wu, Zongze and Zhang, Richard and Shechtman, Eli and Xie, Saining},
  journal={arXiv preprint arXiv:2605.18324},
  year={2026}
}

@inproceedings{zheng2026diffusion,
  title={Diffusion transformers with representation autoencoders},
  author={Zheng, Boyang and Ma, Nanye and Tong, Shengbang and Xie, Saining},
  booktitle={International Conference on Learning Representations (ICLR)},
  year={2026}
}

@inproceedings{yao2025reconstruction,
  title={Reconstruction vs. generation: Taming optimization dilemma in latent diffusion models},
  author={Yao, Jingfeng and Yang, Bin and Wang, Xinggang},
  booktitle={IEEE/CVF Conference on Computer Vision and Pattern Recognition (CVPR)},
  pages={15703--15712},
  year={2025},
  organization={IEEE}
}

@inproceedings{leng2025repa,
  title={{REPA-E}: Unlocking {VAE} for end-to-end tuning with latent diffusion transformers},
  author={Leng, Xingjian and Singh, Jaskirat and Hou, Yunzhong and Xing, Zhenchang and Xie, Saining and Zheng, Liang},
  booktitle={IEEE/CVF International Conference on Computer Vision (ICCV)},
  pages={18262--18272},
  year={2025},
  organization={IEEE}
}

@inproceedings{li2026back,
  title={Back to basics: Let denoising generative models denoise},
  author={Li, Tianhong and He, Kaiming},
  booktitle={Proceedings of the IEEE/CVF Conference on Computer Vision and Pattern Recognition},
  pages={36115--36125},
  year={2026}
}

@inproceedings{shin2026representation,
  title={Representation alignment for {J}ust {I}mage {T}ransformers is not easier than you think},
  author={Shin, Jaeyo and Kim, Jiwook and Shim, Hyunjung},
  booktitle={European Conference on Computer Vision (ECCV)},
  year={2026},
  organization={Springer}
}

@article{chen2025pixelflow,
  title={{P}ixel{F}low: Pixel-space generative models with flow},
  author={Chen, Shoufa and Ge, Chongjian and Zhang, Shilong and Sun, Peize and Luo, Ping},
  journal={arXiv preprint arXiv:2504.07963},
  year={2025}
}

@inproceedings{yu2026pixeldit,
  title={{P}ixel{D}i{T}: Pixel diffusion transformers for image generation},
  author={Yu, Yongsheng and Xiong, Wei and Nie, Weili and Sheng, Yichen and Liu, Shiqiu and Luo, Jiebo},
  booktitle={Proceedings of the IEEE/CVF Conference on Computer Vision and Pattern Recognition},
  pages={14273--14282},
  year={2026}
}

@inproceedings{ma2026deco,
  title={{D}e{C}o: Frequency-decoupled pixel diffusion for end-to-end image generation},
  author={Ma, Zehong and Wei, Longhui and Wang, Shuai and Zhang, Shiliang and Tian, Qi},
  booktitle={Proceedings of the IEEE/CVF Conference on Computer Vision and Pattern Recognition},
  pages={43600--43610},
  year={2026}
}

@article{baade2026latent,
  title={Latent forcing: Reordering the diffusion trajectory for pixel-space image generation},
  author={Baade, Alan and Chan, Eric Ryan and Sargent, Kyle and Chen, Changan and Johnson, Justin and Adeli, Ehsan and Fei-Fei, Li},
  journal={arXiv preprint arXiv:2602.11401},
  year={2026}
}

@article{chen2026asymmetric,
  title={Asymmetric flow models},
  author={Chen, Hansheng and Ackermann, Jan and Kim, Minseo and Wetzstein, Gordon and Guibas, Leonidas},
  journal={arXiv preprint arXiv:2605.12964},
  year={2026}
}

@inproceedings{chen2025masked,
  title={Masked autoencoders are effective tokenizers for diffusion models},
  author={Chen, Hao and Han, Yujin and Chen, Fangyi and Li, Xiang and Wang, Yidong and Wang, Jindong and Wang, Ze and Liu, Zicheng and Zou, Difan and Raj, Bhiksha},
  booktitle={International Conference on Machine Learning (ICML)},
  year={2025}
}

@article{liu2025efficient,
  title={Efficient generative model training via embedded representation warmup},
  author={Liu, Deyuan and Sun, Peng and Li, Xufeng and Lin, Tao},
  journal={arXiv preprint arXiv:2504.10188},
  year={2025}
}

@article{oquab2024dinov2,
  title={{DINOv2}: Learning robust visual features without supervision},
  author={Oquab, Maxime and Darcet, Timoth{\'e}e and Moutakanni, Th{\'e}o and Vo, Huy V. and Szafraniec, Marc and Khalidov, Vasil and Fernandez, Pierre and Haziza, Daniel and Massa, Francisco and El-Nouby, Alaaeldin and others},
  journal={Transactions on Machine Learning Research},
  year={2024}
}

@inproceedings{heusel2017gans,
  title={{GAN}s trained by a two time-scale update rule converge to a local {N}ash equilibrium},
  author={Heusel, Martin and Ramsauer, Hubert and Unterthiner, Thomas and Nessler, Bernhard and Hochreiter, Sepp},
  booktitle={Advances in neural information processing systems (NeurIPS)},
  year={2017}
}

@inproceedings{salimans2016improved,
  title={Improved techniques for training {GAN}s},
  author={Salimans, Tim and Goodfellow, Ian and Zaremba, Wojciech and Cheung, Vicki and Radford, Alec and Chen, Xi},
  booktitle={Advances in neural information processing systems (NeurIPS)},
  pages={2234--2242},
  year={2016}
}

@article{ho2022classifier,
  title={Classifier-free diffusion guidance},
  author={Ho, Jonathan and Salimans, Tim},
  journal={arXiv preprint arXiv:2207.12598},
  year={2022}
}

@inproceedings{kynkaanniemi2024applying,
  title={Applying guidance in a limited interval improves sample and distribution quality in diffusion models},
  author={Kynk{\"a}{\"a}nniemi, Tuomas and Aittala, Miika and Karras, Tero and Laine, Samuli and Aila, Timo and Lehtinen, Jaakko},
  booktitle={Advances in neural information processing systems (NeurIPS)},
  pages={122458--122483},
  year={2024}
}

@inproceedings{hoogeboom2025simpler,
  title={Simpler diffusion: 1.5 {FID} on {I}mage{N}et512 with pixel-space diffusion},
  author={Hoogeboom, Emiel and Mensink, Thomas and Heek, Jonathan and Lamerigts, Kay and Gao, Ruiqi and Salimans, Tim},
  booktitle={IEEE/CVF Conference on Computer Vision and Pattern Recognition (CVPR)},
  pages={18062--18071},
  year={2025},
}
\bibliographystyle{iclr2025_conference}

\newpage
\appendix

\section{Theoretical Analysis and Proofs}
\label{app:proofs}

The analysis connects the construction in Section~\ref{sec:method} to its training role. Markov coupling preserves the JiT input distribution and the oracle regression target; conditional averaging then reduces target and gradient variability. We also quantify how teacher error enters the practical objective, explaining the roles of direct image supervision and reliability-gated selection. Appendix~\ref{app:gaussian-coupling} gives an analytical comparison of observation couplings, and Appendix~\ref{app:related-supervision} discusses the relationship to prior work.

\subsection{Markov Coupling and Martingale Consistency}
\label{app:markov-martingale}

\paragraph{Standing assumptions and notation.}
Let $X\in\mathbb R^d$ satisfy $\mathbb E\|X\|_2^2<\infty$, and let
$c$ denote the conditioning information. For deterministic $u\in(0,1)$,
$m_u(z,c)$ is the posterior mean under the corruption
$Z_u=uX+(1-u)\epsilon$, with $\epsilon\sim\mathcal N(0,I_d)$ independent
of $(X,c)$. All primitive Gaussian noise draws below are independent of
$(X,c)$ and of each other. Unless otherwise stated, times are fixed;
when $(t,s)\sim\pi(t,s)=p(t)q(s\mid t)$, the pair is independent of
$(X,c)$ and the noises, and $p(t)$ is the JiT time distribution.
Write $\operatorname{Var}$ for a covariance matrix, equivalently
$\operatorname{Cov}$ in the main text, and $\preceq$ for the
positive-semidefinite order. All conditional identities hold almost surely.
We assume the weighted losses in use are finite. For statements about
population gradients and their covariances, we additionally assume the
relevant gradients are square-integrable and differentiation can be
interchanged with expectation. These requirements matter because
$w(t)=(1-t)^{-2}$ is unbounded near $t=1$.

\paragraph{Well-posedness and preservation of the corruption marginal.}
For $0<t,s<1$, set
\begin{equation}
 a_{t\mid s}=\frac ts,\qquad
 b_{t\mid s}^2=(1-t)^2-\frac{t^2}{s^2}(1-s)^2.
 \label{eq:app-coupling}
\end{equation}
Then $b_{t\mid s}^2\ge0$ if and only if $t\le s$. For such pairs, choose
$b_{t\mid s}\ge0$ and construct
$Z_s=sX+(1-s)\epsilon_s$ and $Z_t=a_{t\mid s}Z_s+b_{t\mid s}\eta$.
This construction has the JiT marginal and satisfies
$X\perp Z_t\mid(Z_s,c)$.
\begin{proof}
Since all four quantities $t,s,1-t,1-s$ are positive,
\[
 b_{t\mid s}^2\ge0
 \quad\Longleftrightarrow\quad s(1-t)\ge t(1-s)
 \quad\Longleftrightarrow\quad t\le s.
\]
Furthermore,
\[
 Z_t=tX+\frac{t(1-s)}s\epsilon_s+b_{t\mid s}\eta.
\]
The two noise terms are independent, centered Gaussian vectors with total
covariance $(1-t)^2I_d$. Thus
$Z_t\mid(X=x,c)\sim\mathcal N(tx,(1-t)^2I_d)$, as required.
Given $(Z_s,c)$, the remaining randomness in $Z_t$ comes only from
$\eta$, which is independent of $(X,Z_s,c)$. This proves the conditional
Markov property in Eq.~\ref{eq:markov}.
\end{proof}

\begin{proof}[Proof of Proposition~\ref{prop:mart}]
The Markov property implies
$\mathbb E[X\mid Z_s,Z_t,c]=m_s(Z_s,c)$. Therefore,
\begin{align}
 \mathbb E[m_s(Z_s,c)\mid Z_t,c]
 &=\mathbb E\!\left[\mathbb E[X\mid Z_s,Z_t,c]\mid Z_t,c\right]
 =m_t(Z_t,c).
 \label{eq:app-tower}
\end{align}
Here the nested sigma-fields are $\sigma(Z_t,c)$ and
$\sigma(Z_s,Z_t,c)$; neither $\sigma(Z_t,c)$ nor $\sigma(Z_s,c)$ alone
is generally contained in the other. For random times, the same argument
is first applied conditional on $(t,s)$, giving Eq.~\ref{eq:mart}, and then averaged
over $s\mid t$. Preservation of the marginal ensures that conditioning on
$s$ does not change the posterior of $X$ given $(Z_t,t,c)$.
\end{proof}

\paragraph{Composition and the meaning of martingale consistency.}
For $0<t<s_1<s_2<1$, generate the chain
$X\to Z_{s_2}\to Z_{s_1}\to Z_t$ using independent transition noises.
The coefficients satisfy
\begin{equation}
 a_{t\mid s_1}a_{s_1\mid s_2}=a_{t\mid s_2},\qquad
 a_{t\mid s_1}^2b_{s_1\mid s_2}^2+b_{t\mid s_1}^2=b_{t\mid s_2}^2.
 \label{eq:app-composition}
\end{equation}
Indeed, the first equality follows by cancellation, and the second follows
by inserting Eq.~\ref{eq:app-coupling} and cancelling the two terms
containing $(1-s_1)^2$. Thus, after marginalizing out $Z_{s_1}$, the law
of $(X,Z_{s_2},Z_t,c)$ agrees with the direct two-time construction.
The same argument gives compatible couplings for any finite set
$0<\tau_1<\cdots<\tau_n<1$.
For this chain, define
$\mathcal F_i=\sigma(c,Z_{\tau_1},\ldots,Z_{\tau_i})$.
Conditional independence gives
$\mathbb E[X\mid\mathcal F_i]=m_{\tau_i}(Z_{\tau_i},c)$.
Consequently these posterior means form a martingale with respect to
$(\mathcal F_i)$ as the time index increases toward cleaner observations.
This is the precise finite-dimensional sense of the consistency property;
it does not assert pointwise equality of the two predictions.

\paragraph{Log-SNR representation.}
Define $\ell_u=2\log(u/(1-u))$ and
$\xi_u=(Z_u-uX)/(1-u)$. The coupling yields
\begin{equation}
 \xi_t=r(t,s)\xi_s+\sqrt{1-r(t,s)^2}\,\eta,\qquad
 r(t,s)=\frac{t(1-s)}{s(1-t)}
       =\exp\!\left(-\frac{\ell_s-\ell_t}{2}\right).
 \label{eq:app-noise-correlation}
\end{equation}
In particular, $\operatorname{Cov}(\xi_t,\xi_s)=r(t,s)I_d$.
This proves Eq.~\ref{eq:corr}; the stated correlation is between the standardized
corruption noises, rather than between the complete observations.

\subsection{Objective Equivalence and Variance Reduction}
\label{app:objective-variance}

For fixed $t<s$, abbreviate
\[
\begin{aligned}
M_s &= m_s(Z_s,c), &
C_t &= \operatorname{Var}(X\mid Z_t,c),\\
V_t(s) &= \operatorname{Var}(M_s\mid Z_t,c), &
D_t(s) &= \mathbb{E}[
\operatorname{Var}(X\mid Z_s,c)\mid Z_t,c].
\end{aligned}
\]
Here $D_t(s)$ is the covariance removed by the oracle target, whereas
$V_t(s)$ is the covariance that remains.

\paragraph{Conditional squared-loss decomposition.}
Let $U=(Z_t,t,c)$ and let $Y$ be a square-integrable target satisfying
$\mathbb E[Y\mid U]=m_t(Z_t,c)$. Since $f_\theta(U)$ is $U$-measurable,
\begin{equation}
 \mathbb E[w(t)\|f_\theta(U)-Y\|_2^2\mid U]
 =w(t)\|f_\theta(U)-m_t(Z_t,c)\|_2^2
  +w(t)\operatorname{tr}\operatorname{Var}(Y\mid U).
 \label{eq:app-loss-decomposition}
\end{equation}
To verify this identity, expand the square around $m_t(Z_t,c)$.
The cross term is zero because
$\mathbb E[Y-m_t(Z_t,c)\mid U]=0$.

\paragraph{Proof of the objective equivalence in Eq.~\ref{eq:equiv}.}
Proposition~\ref{prop:mart} gives the required conditional mean both for $Y=X$ and
for $Y=M_s$, including when the times are sampled from $\pi$.
Applying Eq.~\ref{eq:app-loss-decomposition} first at fixed $(t,s)$ and
then averaging over $\pi$ proves
\begin{align}
 \mathcal L^\star_{\mathrm{JAx}}(\theta)
 &=\mathcal L_{\mathrm{JiT}}(\theta)+K,\qquad
 K=\mathbb E_{\pi}\!\left[w(t)\operatorname{tr}\{V_t(s)-C_t\}\right].
 \label{eq:app-objective-equivalence}
\end{align}
Expectations with subscript $\pi$ also include the data and coupled noises.
The common marginal $p(t)$ and common weight $w(t)$ are essential here.
The constant $K$ is independent of $\theta$, so the two population
objectives have the same parameter-space minimizers and, wherever
differentiable, the same gradients. This statement concerns the oracle
objective in Eq.~\ref{eq:oracle}, before the sample-dependent activation in Eq.~\ref{eq:batched}.

\begin{proof}[Proof of Proposition~\ref{prop:var}: covariance identity and monotonicity]
The conditional law of total covariance, using the nested information
$(Z_t,c)$ and $(Z_s,Z_t,c)$, gives
\begin{align}
 C_t
 &=\mathbb E[\operatorname{Var}(X\mid Z_s,Z_t,c)\mid Z_t,c]
   +\operatorname{Var}(\mathbb E[X\mid Z_s,Z_t,c]\mid Z_t,c)\notag\\
 &=D_t(s)+V_t(s).
 \label{eq:app-total-covariance}
\end{align}
The second equality uses the Markov property in both terms. Since
$D_t(s)\succeq0$, this proves Eq.~\ref{eq:var}. It also gives the explicit constant
\begin{equation}
 K=-\mathbb E_{\pi}\!\left[w(t)\operatorname{tr}D_t(s)\right]
  =-\mathbb E_{\pi}[w(t)\|X-M_s\|_2^2]\le0.
 \label{eq:app-negative-constant}
\end{equation}

For monotonicity, use the compatible chain
$X\to Z_{s_2}\to Z_{s_1}\to Z_t$ from
Eq.~\ref{eq:app-composition}, where $t<s_1<s_2$.
The Markov property and Proposition~\ref{prop:mart} imply
\[
 \mathbb E[M_{s_2}\mid Z_{s_1},Z_t,c]=M_{s_1}.
\]
Applying conditional total covariance to $M_{s_2}$ now gives
\begin{equation}
 V_t(s_2)-V_t(s_1)
 =\mathbb E[\operatorname{Var}(M_{s_2}\mid Z_{s_1},c)\mid Z_t,c]
 \succeq0.
 \label{eq:app-monotonicity}
\end{equation}
All pair marginals in this calculation agree with Eq.~\ref{eq:coupling}.
Therefore $V_t(s)$ is nondecreasing and $D_t(s)=C_t-V_t(s)$ is
nonincreasing in $s$, in the positive-semidefinite order. At fixed $t$,
the same ordering holds with respect to the log-SNR gap.
\end{proof}

\paragraph{Endpoint limits in Proposition~\ref{prop:var}.}
Under the standing second-moment assumption, the precise limits are
\begin{equation}
 \begin{array}{lll}
 s\downarrow t: & V_t(s)\longrightarrow0, & D_t(s)\longrightarrow C_t,\\[2pt]
 s\uparrow1: & V_t(s)\longrightarrow C_t, & D_t(s)\longrightarrow0.
 \end{array}
 \label{eq:app-endpoint-limits}
\end{equation}
These are $L^1$ limits in matrix trace norm under the common student
marginal of $(Z_t,c)$. The corresponding oracle targets converge in
mean square to $m_t(Z_t,c)$ and $X$, respectively, under the pair coupling.
\begin{proof}
For $s\uparrow1$, the posterior mean minimizes squared error among
functions of $(Z_s,c)$. Comparing it with $Z_s/s$ gives
\begin{equation}
 \mathbb E\operatorname{tr}D_t(s)
 =\mathbb E\|X-M_s\|_2^2
 \le\mathbb E\|X-Z_s/s\|_2^2
 =d\left(\frac{1-s}{s}\right)^2\longrightarrow0.
 \label{eq:app-clean-endpoint}
\end{equation}
Because $D_t(s)$ is positive semidefinite, its trace is its trace norm;
Eq.~\ref{eq:app-total-covariance} gives both limits at this endpoint.

For $s\downarrow t$, realize all two-time pairs using the same
$(X,c,\epsilon,\eta)$, by setting
$Z_s=sX+(1-s)\epsilon$ and
$Z_t^{(s)}=(t/s)Z_s+b_{t\mid s}\eta$.
Both observations converge almost surely to $Z_t^0=tX+(1-t)\epsilon$.
For almost every $c$, the Gaussian likelihood gives the continuous version
\begin{equation}
 m_u(z,c)=
 \frac{\int x\exp\{-\|z-ux\|_2^2/[2(1-u)^2]\}\,P_c(dx)}
      {\int   \exp\{-\|z-ux\|_2^2/[2(1-u)^2]\}\,P_c(dx)},
 \qquad 0<u<1,
 \label{eq:app-posterior-continuity}
\end{equation}
where $P_c$ is the conditional law of $X$ given $c$.
The denominator is strictly positive, and dominated convergence applies
to the numerator and denominator because $\int\|x\|_2\,P_c(dx)<\infty$.
Thus $M_s-m_t(Z_t^{(s)},c)\to0$ almost surely.

For completeness, this convergence is also in mean square.
Put $Q=\|X\|_2^2$. Conditional Jensen gives
$\|M_s\|_2^2\le\mathbb E[Q\mid Z_s,c]$, and similarly for
$\|m_t(Z_t^{(s)},c)\|_2^2$.
The family of conditional expectations of a fixed integrable,
nonnegative $Q$ is uniformly integrable: for any sigma-field
$\mathcal G$, $H=\mathbb E[Q\mid\mathcal G]$, and $A=\{H>L\}$,
\[
 \mathbb E[H\mathbf1_A]
 =\mathbb E[Q\mathbf1_A]
 \le\mathbb E[Q\mathbf1_{\{Q>M\}}]+M\,\mathbb E[Q]/L.
\]
Taking first $L\to\infty$ and then $M\to\infty$ proves the assertion
uniformly in $\mathcal G$. Hence the squared differences above are
uniformly integrable and their expectations tend to zero. Finally,
\[
 \mathbb E\operatorname{tr}V_t(s)
 =\mathbb E\|M_s-m_t(Z_t^{(s)},c)\|_2^2\longrightarrow0.
\]
Each $Z_t^{(s)}$ has the same student marginal, so this is the claimed
$L^1$ covariance limit. Eq.~\ref{eq:app-total-covariance} gives
the remaining limit for $D_t(s)$.
\end{proof}

\subsection{An Analytical Comparison of Observation Couplings}
\label{app:gaussian-coupling}

A scalar Gaussian example isolates the role of coupling without teacher approximation or optimization effects. Let $X\sim\mathcal N(0,\tau^2)$, with $\tau>0$, and fix $0<t<s<1$. Construct $Z_u=uX+(1-u)\epsilon_u$ for $u\in\{t,s\}$, where $(\epsilon_t,\epsilon_s)$ is jointly Gaussian, independent of $X$, with standard normal marginals and correlation $\rho$. All choices of $\rho$ preserve the individual corruption marginals. Define
\begin{equation}
 Q_u=u^2\tau^2+(1-u)^2,\qquad
 C_{ts}=ts\tau^2+(1-t)(1-s)\rho,\qquad
 m_u(z)=\frac{u\tau^2}{Q_u}z.
 \label{eq:app-gaussian-posterior}
\end{equation}
Gaussian conditioning gives the oracle target's conditional mean and variance:
\begin{equation}
 \mathbb E[m_s(Z_s)\mid Z_t=z]=\frac{s\tau^2C_{ts}}{Q_sQ_t}z,\qquad
 \operatorname{Var}(m_s(Z_s)\mid Z_t)
 =\left(\frac{s\tau^2}{Q_s}\right)^2
  \left(Q_s-\frac{C_{ts}^2}{Q_t}\right).
 \label{eq:app-gaussian-target}
\end{equation}
Subtracting $m_t(z)$ from the conditional mean yields
\begin{equation}
 \mathbb E[m_s(Z_s)\mid Z_t=z]-m_t(z)
 =\frac{\tau^2(1-s)\{s(1-t)\rho-t(1-s)\}}{Q_sQ_t}z.
 \label{eq:app-gaussian-bias}
\end{equation}
Thus, within this family, conditional consistency holds precisely at
$\rho=t(1-s)/(s(1-t))$, the Markov correlation in Eq.~\ref{eq:corr}.
Independent noise ($\rho=0$) and shared noise ($\rho=1$) both shift the conditional mean for nonzero $z$.

\begin{table}[t]
    \centering
    \small
    \setlength{\tabcolsep}{6pt}
    \renewcommand{\arraystretch}{1.10}
    \caption{Analytical target statistics for $X\sim\mathcal N(0,1)$, $t=1/3$, and $s=1/2$. The required conditional mean is $m_t(z)=0.6z$. These values are exact Gaussian calculations.}
    \label{tab:gaussian-couplings}
    \begin{tabular*}{\linewidth}{@{\extracolsep{\fill}}|l|c|c|c|}
        \hline
        \rowcolor{gray!15}
        Target / coupling & $\rho$ & $\mathbb E[Y\mid Z_t=z]$ & $\operatorname{Var}(Y\mid Z_t)$ \\
        \hline
        Clean image $X$ & -- & $0.6z$ & $0.80$ \\
        Oracle, independent noise & $0$ & $0.3z$ & $0.45$ \\
        Oracle, shared noise & $1$ & $0.9z$ & $0.05$ \\
        \rowcolor{gray!8}
        Oracle, Markov coupling & $0.5$ & $0.6z$ & $0.30$ \\
        \hline
    \end{tabular*}
\end{table}

Table~\ref{tab:gaussian-couplings} illustrates why the two target criteria in Section~\ref{sec:target_requirements} are needed together. Shared-noise supervision has the smallest conditional variance in this example, but changes the conditional regression target. Markov coupling preserves that target while reducing its variance relative to clean-image supervision. This provides a concrete explanation for the coupling design; the JiT ablation in Table~\ref{tab:pair_coupling_ablation} evaluates its practical effect with an EMA teacher.

\subsection{Implications for Stochastic Gradients}
\label{app:stochastic-gradients}

Fix the current student parameters $\theta$, and hold every target fixed
when differentiating. Let $U=(Z_t,t,c)$ and
$J_\theta(U)=\partial f_\theta(U)/\partial\theta\in\mathbb R^{d\times p}$,
where $p$ is the number of parameters. For a stop-gradient target $Y$,
\begin{equation}
 g(Y)=2w(t)J_\theta(U)^\top(f_\theta(U)-Y).
 \label{eq:app-sample-gradient}
\end{equation}
We use this expression to translate the target covariance reduction in
Proposition~\ref{prop:var} into a corresponding reduction in stochastic-gradient
covariance, and then relate it to the normalized gradient statistic evaluated
in Figure~\ref{fig:gradient_variance}.

At fixed $(t,s)$, the matrix $J_\theta(U)$ is fixed conditional on $U$.
Thus, writing $J=J_\theta(U)$ and $w=w(t)$,
\begin{align}
 \mathbb E[g(M_s)\mid U]&=\mathbb E[g(X)\mid U]
 =2wJ^\top(f_\theta(U)-m_t(Z_t,c)),\notag\\
 \operatorname{Var}(g(Y)\mid U)&=
 4w^2J^\top\operatorname{Var}(Y\mid U)J.
 \label{eq:app-gradient-moments}
\end{align}
Consequently, Proposition~\ref{prop:var} gives
\begin{equation}
 \operatorname{Var}(g(X)\mid U)-\operatorname{Var}(g(M_s)\mid U)
 =4w^2J^\top D_t(s)J\succeq0.
 \label{eq:app-gradient-order}
\end{equation}
Thus the oracle prediction target preserves the expected gradient while
reducing its conditional covariance. For random pairs, the same conclusion
follows by conditioning first on $(U,s)$ and then on $U$, since the conditional
mean above does not depend on $s$. The corresponding unconditional ordering
then follows from the law of total covariance.

\paragraph{The normalized combined oracle gradient.}
Figure~\ref{fig:gradient_variance} compares the combined objective with JiT
after dividing its gradient by $1+\lambda$. For an active pair, the
corresponding oracle quantity is
\begin{equation}
 G_\lambda^\star=
 \frac{g(X)+\lambda g(M_s)}{1+\lambda},
 \qquad \lambda\ge0.
 \label{eq:app-combined-gradient}
\end{equation}
The two gradients share $(X,Z_s,Z_t)$ and are generally correlated.
Set $R=X-M_s$. Since
$\mathbb E[R\mid Z_s,Z_t,c]=0$, $R$ is conditionally orthogonal to the
oracle prediction fluctuation $M_s-m_t(Z_t,c)$, and
$\mathbb E[RR^\top\mid U]=D_t(s)$. It follows that
\begin{align}
 \operatorname{Var}(G_\lambda^\star\mid U)
 &=4w^2J^\top
 \left[V_t(s)+\frac{D_t(s)}{(1+\lambda)^2}\right]J,
 \label{eq:app-combined-covariance}\\
 \operatorname{Var}(g(X)\mid U)
 -\operatorname{Var}(G_\lambda^\star\mid U)
 &=4w^2\left(1-\frac1{(1+\lambda)^2}\right)
 J^\top D_t(s)J\succeq0.
 \label{eq:app-combined-reduction}
\end{align}
The identity follows from
$g(X)=g(M_s)-2wJ^\top(X-M_s)$
and the conditional orthogonality above.
The normalized combined oracle gradient therefore has the same conditional
mean as the JiT gradient and no larger conditional covariance. The same
conclusion extends to random pairs and to unconditional covariances as above.
For a mean of $B$ independent, identically distributed sample gradients
at fixed parameters $\theta$, each covariance is divided by $B$. No such variance ordering
against JiT is implied for the \emph{unnormalized} combined gradient.

\paragraph{Empirical evidence with an EMA teacher.}
Figure~\ref{fig:gradient_variance} tests the same variance-reduction mechanism
using a frozen EMA teacher at the evaluated student checkpoints.
Normalizing the combined gradient by $1+\lambda$ removes the overall scale
difference. The resulting $9.6\%$ average variance reduction is accompanied
by a $6.5\%$ reduction in mean-squared deviation from the corresponding JiT
population gradient (averaged over epoch 200 and 600). The latter accounts for both gradient variance and a
possible mean-gradient shift, so the observed reduction is not explained by
variance alone.

The oracle analysis above identifies the gradient-level mechanism induced by
the Markov-coupled prediction target. The EMA measurements exhibit the same
variance-reduction behavior in the evaluated practical training setting.
Appendix~\ref{app:ema-reliability} further characterizes how teacher
approximation error enters this objective and how reliability gating and
direct JiT supervision control its effect.

\subsection{EMA Approximation and Reliability Gating}
\label{app:ema-reliability}

We now characterize the practical construction used by \ourwork, including
the EMA teacher, reliability-gated pair selection, and direct clean-image
supervision.

Condition throughout this subsection on the training history
$\mathcal H_k$ before the current batch. The student parameters, EMA
parameters, and risk estimates based on previous updates are then fixed.
Assume the current data and corruption noises are freshly drawn from the
specified training distribution, and that pair selection uses only the
student time and $\mathcal H_k$, as in Section~\ref{sec:band}. All
expectations below are conditional on $\mathcal H_k$, which is suppressed
in the notation. The teacher and selection rule are held fixed when
differentiating the current loss; no derivative is taken through their
training history.

\paragraph{Feasibility and the nearest admissible teacher.}
Assume $0<\Delta_{\min}\le\Delta_{\max}<\infty$ and
$\ell_{\max}<\infty$. Given a finite frontier $\ell_{\mathrm{safe}}(k)$,
define the frontier-based admissible set
\begin{equation}
 \mathcal I_k(t)=
 [\ell_t+\Delta_{\min},\ell_t+\Delta_{\max}]
 \cap[\ell_{\mathrm{safe}}(k),\ell_{\max}].
 \label{eq:app-admissible-set}
\end{equation}
Put
\[
 L_k(t)=\max\{\ell_t+\Delta_{\min},\ell_{\mathrm{safe}}(k)\},
 \qquad
 U_k(t)=\min\{\ell_t+\Delta_{\max},\ell_{\max}\}.
\]
Then $\mathcal I_k(t)$ is nonempty if and only if
$L_k(t)\le U_k(t)$, which is exactly the activation test in
Section~\ref{sec:band}. Whenever active,
$\ell_s^\star=L_k(t)$ is its smallest element and
$s^\star=\operatorname{sigmoid}(\ell_s^\star/2)$ satisfies
$t<s^\star<1$. By Eq.~\ref{eq:app-monotonicity}, this choice maximizes
the oracle covariance reduction $D_t(s)$ within $\mathcal I_k(t)$.
If the frontier is unavailable or the set is empty, define
$a_k(t)=0$; otherwise set $a_k(t)=1$.

The frontier is defined from the empirical prediction-risk curve and serves
as the practical reliability criterion. For a nonmonotone empirical curve
$R_k(\ell)$, the definition
\[
 \ell_{\mathrm{safe}}(k)
 =
 \inf\{\ell:R_k(\ell)\le R^\star\}
 +\delta_{\mathrm{safe}}
\]
does not by itself imply $R_k(\ell)\le R^\star$ for every
$\ell\ge\ell_{\mathrm{safe}}(k)$. Such a statement would require additional
regularity of the risk curve together with an estimation guarantee.
The margin $\delta_{\mathrm{safe}}$ is therefore a design margin rather than
a finite-sample confidence bound.

\paragraph{Conditional bias of an imperfect teacher.}
For an active pair, write
\[
 \bar x_s=M_s+e_s,\qquad
 e_s=f_{\bar\theta}(Z_s,s,c)-m_s(Z_s,c),\qquad
 \beta_{t,s}(U)=\mathbb E[e_s\mid U].
\]
Here $s=s^\star(t)$ is fixed given the student time and history.
Proposition~\ref{prop:mart} gives
\[
 \mathbb E[\bar x_s\mid U]
 =
 m_t(Z_t,c)+\beta_{t,s}(U).
\]
Let $h=f_\theta(U)-m_t(Z_t,c)$. Direct expansion, retaining the
conditional expectation in every random term, yields
\begin{align}
 \mathbb E[w\|f_\theta(U)-\bar x_s\|_2^2\mid U]
 ={}&w\|h\|_2^2
 -2w\langle h,\beta_{t,s}(U)\rangle \notag\\
 &+w\mathbb E[
 \|\bar x_s-m_t(Z_t,c)\|_2^2\mid U].
 \label{eq:app-biased-loss}
\end{align}
The last term is independent of the current student parameters under the
frozen-teacher convention. Hence
\begin{equation}
 \mathbb E[g(\bar x_s)-g(X)\mid U]
 =
 -2wJ^\top\beta_{t,s}(U).
 \label{eq:app-ema-gradient-bias}
\end{equation}
Thus the conditional mean of the teacher error directly determines the shift
in the expected gradient.

Teacher approximation also enters the target covariance:
\begin{align}
 \operatorname{Var}(\bar x_s\mid U)
 ={}&V_t(s)+\operatorname{Var}(e_s\mid U)
   +\operatorname{Cov}(M_s,e_s\mid U) \notag\\
 &+\operatorname{Cov}(e_s,M_s\mid U).
 \label{eq:app-ema-covariance}
\end{align}
This decomposition separates the covariance of the oracle target from the
additional teacher-error terms. It makes explicit the two aspects controlled
by the practical construction: the conditional mean shift is governed by
$\beta_{t,s}(U)$, while the remaining variability depends on the magnitude
and correlation structure of the teacher error.

\paragraph{Prediction risk as a measure of teacher reliability.}
Clean-image prediction error provides an empirically estimable proxy for
teacher approximation error. The Markov property gives
$\mathbb E[X-M_s\mid Z_s,Z_t,c]=0$, while $e_s$ is measurable with respect
to $(Z_s,c)$. Consequently,
\begin{equation}
 \mathbb E[\|\bar x_s-X\|_2^2\mid U]
 =
 \mathbb E[\|e_s\|_2^2\mid U]
 +\operatorname{tr}D_t(s),
 \qquad
 \|\beta_{t,s}(U)\|_2^2
 \le
 \mathbb E[\|e_s\|_2^2\mid U].
 \label{eq:app-risk-bias}
\end{equation}
Define the population prediction risk at
$s=\operatorname{sigmoid}(\ell/2)$ by
\[
 \mathcal R_k(\ell)
 =
 \mathbb E\|f_{\bar\theta}(Z_s,s,c)-X\|_2^2.
\]
For each fixed selected pair,
\begin{equation}
 \mathbb E\|\beta_{t,s}(U)\|_2^2
 \le
 \mathbb E\|e_s\|_2^2
 =
 \mathcal R_k(\ell_s)
 -\mathbb E\|X-M_s\|_2^2
 \le
 \mathcal R_k(\ell_s).
 \label{eq:app-population-risk-bound}
\end{equation}
Thus small clean-image prediction risk controls the average squared
conditional bias of the teacher target, providing a direct motivation for
the reliability criterion used in Section~\ref{sec:band}.

The bound concerns the population risk and controls the bias on average,
rather than pointwise for every input. The online quantity $R_k$ estimates
$\mathcal R_k$ and should not be substituted for it in an exact inequality
without an estimation guarantee. If the implemented risk is averaged over
the $d$ coordinates, the corresponding population quantity above is scaled
by $d$. A numerical gradient-bias bound additionally depends on $w(t)$ and
the network Jacobian, as indicated by
Eq.~\ref{eq:app-ema-gradient-bias}.

\paragraph{The mixed target and time weighting.}
For a fixed teacher and selection rule, let
$q_k(t)=\lambda a_k(t)$. On an active pair, completing the square gives
\begin{align}
 Y_{\mathrm{mix},k}
 &=
 \frac{X+q_k(t)\bar x_{s^\star(t)}}{1+q_k(t)},
 \label{eq:app-mixed-target}\\
 \ell_{\mathrm{total},k}
 &=
 w(t)\|f_\theta(U)-X\|_2^2
 +
 w(t)q_k(t)
 \|f_\theta(U)-\bar x_{s^\star(t)}\|_2^2\notag\\
 &=
 w(t)(1+q_k(t))
 \|f_\theta(U)-Y_{\mathrm{mix},k}\|_2^2
 +
 \frac{w(t)q_k(t)}{1+q_k(t)}
 \|X-\bar x_{s^\star(t)}\|_2^2.
 \label{eq:app-mixed-loss}
\end{align}
The final term is independent of $\theta$ under the frozen-teacher
convention. For inactive times, set $Y_{\mathrm{mix},k}=X$ and the final
term to zero, without evaluating a teacher.

The alignment therefore changes both the regression target and the effective
time weight, from $w(t)$ to $w(t)(1+\lambda a_k(t))$. This identity motivates
the reweighting-only control in Table~\ref{tab:pair_coupling_ablation}:
replaying the same activation schedule while regressing only to $X$ isolates
the contribution of the teacher target from the induced change in weighting.

\paragraph{Bias--variance trade-off with an imperfect teacher.}
For an active pair, let
$\alpha=\lambda/(1+\lambda)$, so that
$Y_{\mathrm{mix},k}=(1-\alpha)X+\alpha(M_s+e_s)$.
Write $m_t=m_t(Z_t,c)$ and define
\[
 v=\operatorname{tr}V_t(s),\qquad
 \delta=\operatorname{tr}D_t(s),\qquad
 e_2=\mathbb E[\|e_s\|_2^2\mid U],\qquad
 b=\mathbb E[\langle M_s-m_t,e_s\rangle\mid U].
\]
Since $X-M_s$ has zero conditional mean given $(Z_s,Z_t,c)$,
it is conditionally orthogonal to both $M_s-m_t$ and $e_s$.
Expanding the mixed target's error gives the exact identity
\begin{equation}
 \mathbb E[
 \|Y_{\mathrm{mix},k}-m_t\|_2^2
 -
 \|X-m_t\|_2^2
 \mid U]
 =
 -\alpha(2-\alpha)\delta
 +2\alpha b
 +\alpha^2e_2.
 \label{eq:app-mixed-error-budget}
\end{equation}
The first term is the oracle variance benefit, while the remaining terms
quantify teacher approximation error and its correlation with the oracle
prediction. This decomposition makes explicit the trade-off underlying the
practical rule: nearby states provide a larger oracle variance reduction,
whereas prediction accuracy controls the additional approximation terms.

By $|b|\le\sqrt{ve_2}$, a sufficient condition for the mixed target to have
no greater conditional squared error around $m_t$ than $X$ is
\begin{equation}
 2\sqrt{ve_2}+\alpha e_2
 \le
 (2-\alpha)\delta.
 \label{eq:app-error-budget-condition}
\end{equation}
This condition is sufficient rather than necessary and is not used directly
by the pair-selection rule. It illustrates one regime in which the oracle
variance benefit dominates teacher approximation error. The same
decomposition applies to conditional gradient deviation by using
$\|J^\top z\|_2^2$ in place of $\|z\|_2^2$ and multiplying by $4w^2$.

\paragraph{The actual gated objective and its minimizer.}
Since Eq.~\ref{eq:batched} divides by the full batch size, its conditional
population counterpart is
\begin{equation}
 \mathcal L_{\mathrm{align},k}(\theta)
 =
 \mathbb E[
 a_k(t)w(t)
 \|f_\theta(U)-\bar x_{s^\star(t)}\|_2^2].
 \label{eq:app-gated-objective}
\end{equation}
For inactive times the integrand is defined to be zero without requiring
a teacher state. Set
$\beta_k(U)=\beta_{t,s^\star(t)}(U)$ on active times and zero otherwise.
Combining Eq.~\ref{eq:app-biased-loss} with the JiT term gives, up to a term
independent of $\theta$,
\begin{align}
 &\mathcal L_{\mathrm{JiT}}
 +\lambda\mathcal L_{\mathrm{align},k}\notag\\
 &\quad=
 \mathbb E\!\left[
 w(t)(1+\lambda a_k(t))
 \|f_\theta(U)-m_t(Z_t,c)\|_2^2
 \right]\notag\\
 &\qquad
 -2\lambda
 \mathbb E\!\left[
 a_k(t)w(t)
 \langle
 f_\theta(U)-m_t(Z_t,c),
 \beta_k(U)
 \rangle
 \right]
 +\mathrm{const}.
 \label{eq:app-gated-total}
\end{align}
Completing the square shows that the pointwise optimal predictor over
unrestricted square-integrable functions is
\begin{equation}
 f_k^{\mathrm{opt}}(U)
 =
 m_t(Z_t,c)
 +
 \frac{\lambda a_k(t)}
      {1+\lambda a_k(t)}
 \,\beta_k(U).
 \label{eq:app-shifted-optimum}
\end{equation}
This identity makes the role of direct JiT supervision explicit. On an active
pair, the conditional mean shift is attenuated from $\beta_k(U)$ under
teacher-only regression to
\[
 \frac{\lambda}{1+\lambda}\beta_k(U)
\]
under the combined objective. For the experimental setting $\lambda=0.5$,
the attenuation factor is $1/3$, so the shift has one third of the
teacher-only magnitude and one ninth of its squared magnitude.

The expression above describes the frozen-teacher conditional regression
optimum; the trained network additionally reflects its function class and
optimization. Even with an oracle teacher $(\beta_k=0)$, gating changes the
quadratic weight from $w(t)$ to $w(t)(1+\lambda a_k(t))$. The unrestricted
Bayes predictor remains $m_t$, while in a restricted network class the
parameter-space optimum can depend on this weighting. Without gating and
with an oracle teacher, Eq.~\ref{eq:app-objective-equivalence} recovers the
exact identity
\[
 \mathcal L_{\mathrm{JiT}}
 +
 \lambda\mathcal L^\star_{\mathrm{JAx}}
 =
 (1+\lambda)\mathcal L_{\mathrm{JiT}}
 +
 \lambda K.
\]

Together, these results clarify the complementary roles of the three design
choices. Markov coupling supplies the conditionally consistent low-variance
target, reliability-gated pair selection balances the oracle preference for
nearby states with empirical teacher accuracy, and direct clean-image
supervision attenuates residual teacher bias. The practical EMA construction
therefore follows the same prediction-supervision principle identified by
the oracle analysis.

\section{Relation to Existing Supervision Methods}
\label{app:related-supervision}
\begin{figure}[h]
    \centering
    \includegraphics[width=\linewidth]{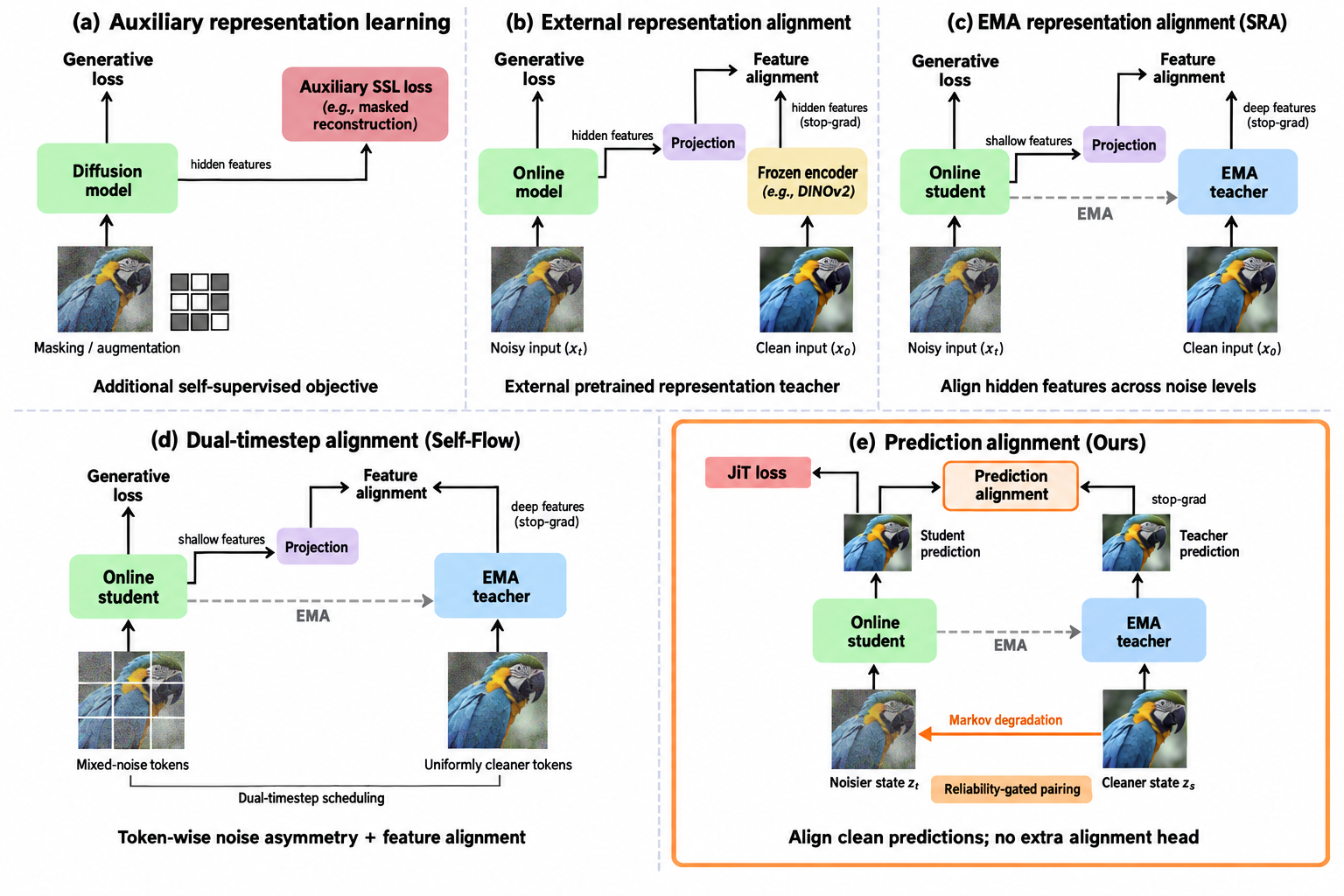}
    \caption{\textbf{Comparison of auxiliary supervision strategies
for generative modeling.}
(a) Auxiliary self-supervised learning, such as masked reconstruction.
(b) Hidden-feature alignment with a frozen pretrained encoder.
(c) SRA: alignment between shallow student and deeper EMA-teacher
features.
(d) Self-Flow: feature alignment with dual-timestep noise scheduling.
(e) \ourwork{}: direct clean-prediction alignment under
reliability-gated Markov coupling, without an additional
trainable alignment head.
All methods retain the base generative objective.
Dashed arrows indicate EMA updates, and teacher targets are
stop-gradient.
Inputs are schematic; conditioning, latent encoding, and
ground-truth target connections are omitted for clarity.}
    \label{fig:alignment_paradigms}
\end{figure}

\paragraph{Comparison with representation-based supervision.}
Figure~\ref{fig:alignment_paradigms} compares \ourwork{} with
four representative strategies for augmenting generative training.
Auxiliary representation learning introduces self-supervised tasks,
such as masked reconstruction, while external representation
alignment uses a pretrained encoder to supervise hidden features.
SRA replaces the external encoder with an EMA teacher and aligns
shallow student features with deeper teacher features.
Self-Flow further introduces dual-timestep scheduling, using
mixed-noise student inputs and uniformly cleaner teacher inputs
for feature alignment.
\ourwork{} instead applies alignment directly to clean predictions,
using reliability-gated Markov coupling to construct teacher--student
pairs.
This formulation requires neither an external representation encoder
nor an additional trainable alignment head, and avoids imposing
an auxiliary matching constraint on hidden representations.
Since prediction alignment and representation alignment act at
different levels, they are complementary and can be combined.

\paragraph{Stochastic and trajectory consistency.}
Consistent Diffusion Models~\citep{daras2023consistent} enforce
a martingale property under the model's learned reverse process
to mitigate sampling drift.
\ourwork{} constructs teacher--student observations directly
through an explicit forward Markov coupling
$X \to Z_s \to Z_t$, preserving the JiT corruption marginals
without simulating model-generated trajectories.
This coupling makes the cleaner oracle prediction a valid
regression target for the noisier student:
its conditional expectation equals the student's desired
posterior mean.
The resulting oracle alignment objective therefore has the
same population gradient as JiT, with reduced conditional
target covariance.
Consistency Models~\citep{song2023consistency} instead learn
a common endpoint prediction along each probability-flow
ODE trajectory for one- or few-step generation.
Our stochastic coupling permits oracle predictions to differ
across individual pairs while preserving their conditional
expectation.
This distinction enables auxiliary prediction supervision
within the existing JiT training and sampling framework.

\paragraph{Low-variance prediction targets.}
Stable Velocity~\citep{yang2026stable} introduces StableVM,
which reduces target variance through a weighted aggregation
of conditional velocities over multiple reference data points.
\ourwork{} obtains its oracle target by posterior conditioning
on a coupled, cleaner observation of the same image.
The Markov construction jointly establishes conditional-mean
correctness and identifies the exact covariance removed:
the expected posterior uncertainty remaining at the teacher
observation.
Both properties matter, since a low-variance target can still
supervise the wrong conditional mean, as illustrated by the
Gaussian example in Appendix~\ref{app:gaussian-coupling}.
The analysis also informs teacher-time selection:
closer oracle targets provide greater variance reduction,
whereas practical EMA targets must be sufficiently reliable.
Our reliability-gated coupling band operationalizes this
trade-off using an empirical reliability frontier and bounded
log-SNR separation, retaining JiT-only supervision when no
feasible pair exists.
The contribution thus connects an explicit coupling,
a posterior-based target construction, and reliability-aware
pair selection into a practical prediction-alignment method.

\section{Implementation Details}
\label{app:implementation}

Algorithm~\ref{alg:jax} summarizes a training update. Table~\ref{tab:implementation} summarizes the backbone architectures, training configurations, alignment settings, and evaluation protocol.

\paragraph{JiT-S/4.}
For the ImageNet $32\times32$ experiments in Figure~\ref{fig:imagenet32}, we construct JiT-S/4 from JiT-B by halving the hidden dimension from 768 to 384 and using $4\times4$ patches.

\begin{algorithm}[t]
    \caption{One training update of~\ourwork}
    \label{alg:jax}
    \begin{algorithmic}[1]
        \REQUIRE Student $\theta$, EMA teacher $\bar\theta$, EMA decay $\rho$, previous risk estimates $R_k$
        \STATE Draw a batch $\{(X_i,c_i)\}_{i=1}^B$ and student times $t_i\sim p(t)$.
        \STATE Determine activation using Eqs.~\ref{eq:frontier}--\ref{eq:teacher_selection} and the feasibility conditions; set $s_i^\star=s^\star_k(t_i)$ for active samples.
        \FOR{$i=1,\ldots,B$}
            \IF{$a_k(t_i)=1$}
                \STATE Draw $(Z_{s_i^\star},Z_{t_i})$ using Eq.~\ref{eq:coupling}; set $\bar{x}_{s_i^\star}=\mathrm{sg}[f_{\bar\theta}(Z_{s_i^\star},s_i^\star,c_i)]$.
            \ELSE
                \STATE Draw $Z_{t_i}=t_iX_i+(1-t_i)\epsilon_i$, with $\epsilon_i\sim\mathcal{N}(0,I)$.
            \ENDIF
        \ENDFOR
        \STATE Update $\theta$ using the batch JiT loss plus $\lambda\widehat{\mathcal{L}}_{\mathrm{JA}x,k}$.
        \STATE Update $\bar\theta\leftarrow\rho\bar\theta+(1-\rho)\theta$.
        \STATE At calibration intervals, refresh the risk estimates for subsequent updates.
    \end{algorithmic}
\end{algorithm}


\section{Limitation and Future Work}
Our evaluation focuses on class-conditional image generation, leaving large-scale text-to-image and video generation unexplored. We expect \ourwork{} to be effective in these settings as well, since its prediction alignment mechanism does not rely on an external pretrained representation encoder. This property may be particularly valuable for video generation, where higher-dimensional spatiotemporal representations can intensify competition for model capacity between representation alignment and denoising. By aligning clean predictions directly, \ourwork{} avoids introducing an additional hidden-representation matching objective. Moreover, \ourwork{} is complementary to hidden-representation alignment methods and can be used together with them, offering a promising direction for future work. Our reliability frontier and several hyperparameter choices are empirically determined. In the future, we plan to systematically study how to better translate the benefits of oracle targets to practical EMA teachers and develop more principled strategies for hyperparameter selection.

\begin{table}[h]
    \centering
    \caption{Implementation details of \ourwork on ImageNet $256\times256$. The backbone architectures and base training configurations follow JiT. The alignment weight is shared across model scales. Training-teacher EMA settings are listed separately from the EMA candidates used for generation evaluation.}
    \label{tab:implementation}
    \small
    \renewcommand{\arraystretch}{1.08}
    \setlength{\tabcolsep}{6pt}
    \begin{tabular*}{\textwidth}{@{\extracolsep{\fill}}|l|ccc|}
        \hline
        \rowcolor{gray!15}
        Configuration & \ourwork-B/16 & \ourwork-L/16 & \ourwork-H/16 \\
        \hline

        \rowcolor{gray!8}
        \multicolumn{4}{|l|}{\textbf{Architecture}} \\
        \hline
        Depth & 12 & 24 & 32 \\
        Hidden dimension & 768 & 1024 & 1280 \\
        Attention heads & 12 & 16 & 16 \\
        Image size & \multicolumn{3}{c|}{$256\times256$} \\
        Patch size & \multicolumn{3}{c|}{$16\times16$} \\
        Bottleneck dimension & 128 & 128 & 256 \\
        Dropout & 0 & 0 & 0.2 \\
        In-context class tokens & \multicolumn{3}{c|}{32} \\
        In-context start block & 4 & 8 & 10 \\
        Prediction parameterization & \multicolumn{3}{c|}{$\bm{x}$-prediction} \\
        \hline

        \rowcolor{gray!8}
        \multicolumn{4}{|l|}{\textbf{Base training}} \\
        \hline
        Training epochs & \multicolumn{3}{c|}{600} \\
        Training hardware & \multicolumn{3}{c|}{8 NVIDIA RTX Pro 6000 or 8 NVIDIA A100 GPUs} \\
        Optimizer & \multicolumn{3}{c|}{AdamW, $(\beta_1,\beta_2)=(0.9,0.95)$} \\
        Batch size & \multicolumn{3}{c|}{1024} \\
        Learning rate & \multicolumn{3}{c|}{$2\times10^{-4}$} \\
        Learning-rate schedule & \multicolumn{3}{c|}{Constant after warmup} \\
        Warmup epochs & \multicolumn{3}{c|}{5} \\
        Weight decay & \multicolumn{3}{c|}{0} \\
        Student time distribution & \multicolumn{3}{c|}{$\operatorname{logit}(t)\sim\mathcal{N}(-0.8,0.8^2)$} \\
        Noise scale & \multicolumn{3}{c|}{1.0} \\
        Minimum division denominator & \multicolumn{3}{c|}{0.05} \\
        Class-conditioning dropout & \multicolumn{3}{c|}{0.1} \\
        \hline

        \rowcolor{gray!8}
        \multicolumn{4}{|l|}{\textbf{Prediction alignment and pair selection}} \\
        \hline
        Alignment weight $\lambda$ & \multicolumn{3}{c|}{0.5} \\
        Teacher EMA decay & \multicolumn{3}{c|}{0.9996} \\
        Teacher initialization & \multicolumn{3}{c|}{Copy of the initial student parameters} \\
        Observation coupling & \multicolumn{3}{c|}{Markov coupling, Eq.~\ref{eq:coupling}} \\
        Minimum log-SNR gap $\Delta_{\min}$ & \multicolumn{3}{c|}{2.0} \\
        Maximum log-SNR gap $\Delta_{\max}$ &  \multicolumn{3}{c|}{2.5} \\
        Maximum teacher log-SNR $\ell_{\max}$ & \multicolumn{3}{c|}{$2\log(0.95/0.05)\approx5.889$} \\
        Admissible prediction risk $R^\star$ & \multicolumn{3}{c|}{0.0155} \\
        Safety margin $\delta_{\mathrm{safe}}$ & \multicolumn{3}{c|}{0.25} \\
        Risk-estimation grid spacing & \multicolumn{3}{c|}{0.25} \\
        Calibration update interval & \multicolumn{3}{c|}{Every epoch} \\
        Alignment normalization & \multicolumn{3}{c|}{Full batch size $B$} \\
        Infeasible pairs & \multicolumn{3}{c|}{JiT supervision only} \\
        \hline

        \rowcolor{gray!8}
        \multicolumn{4}{|l|}{\textbf{Sampling and evaluation}} \\
        \hline
        ODE solver & \multicolumn{3}{c|}{Heun} \\
        ODE steps & \multicolumn{3}{c|}{50} \\
        Sampling timesteps & \multicolumn{3}{c|}{Uniformly spaced in $[0,1]$} \\
        CFG interval & \multicolumn{3}{c|}{$[0.1,1.0]$} \\
        Final CFG candidates (600 epochs) & \multicolumn{3}{c|}{$1.0-4.0$} \\
        Evaluation EMA candidates & \multicolumn{3}{c|}{$\{0.9996,0.9998,0.9999\}$} \\
        Samples for CFG/EMA selection & \multicolumn{3}{c}{10,000 for sweeping; 50,000 for final evaluation}\\
        Generation batch size per GPU & \multicolumn{3}{c|}{128} \\
        Samples for final evaluation & \multicolumn{3}{c|}{50,000} \\
        Evaluation metrics & \multicolumn{3}{c|}{FID $\downarrow$, IS $\uparrow$} \\
        \hline
    \end{tabular*}
\end{table}

\section{Generation Quality, Convergence, and Training Cost}
\label{app:metrics-cost}

\paragraph{Sustained FID improvements across scales.}
Table~\ref{tab:jit_scales} reports results on ImageNet
$256\times256$ after 600 training epochs.
With unchanged student architectures and sampling procedures,
\ourwork{} reduces FID from 3.66 to 3.41 on B/16,
from 2.36 to 2.26 on L/16, and from 1.86 to 1.82 on H/16.
Notably, \ourwork{}-H/16 matches the FID of the larger JiT-G/16
model (1.82), using approximately half as many parameters
(953M versus 2B).

\begin{table}[h]
    \centering
    \small
    \setlength{\tabcolsep}{4pt}
    \renewcommand{\arraystretch}{1.10}
    \caption{Results on ImageNet $256\times256$ after 600 epochs,
    with unchanged JiT architectures and sampling procedures.
    Bold marks the lower FID at each scale.}
    \label{tab:jit_scales}
    \begin{tabular*}{\linewidth}{@{\extracolsep{\fill}}|l|rr|rr|}
        \hline
        \rowcolor{gray!15}
        Scale & \multicolumn{2}{c|}{FID $\downarrow$}
              & \multicolumn{2}{c|}{IS $\uparrow$} \\
        \rowcolor{gray!15}
        & JiT & \ourwork{} & JiT & \ourwork{} \\
        \hline
        B/16 & 3.66 & \textbf{3.41} & 275.1 & 271.82 \\
        L/16 & 2.36 & \textbf{2.26} & 298.5 & 284.93 \\
        H/16 & 1.86 & \textbf{1.82} & 303.4 & 302.8 \\
        \hline
    \end{tabular*}
\end{table}

\paragraph{Accounting for the practical overhead.}
The additional training workload consists of stop-gradient
teacher evaluation on active samples, elementwise EMA updates,
and periodic risk calibration.
The teacher workload scales with the active fraction,
while full-batch normalization keeps the loss weight of each
active target fixed.
The student architecture and sampling procedure remain unchanged,
so \ourwork{} introduces no additional inference cost.

For an approximate compute accounting, let $F$ denote the cost
of a full-batch student forward pass and count a backward pass
as two forward passes.
A baseline JiT step then costs approximately $3F$.
Evaluating the teacher on an active fraction $\alpha$ adds
$\alpha F$, giving a relative cost of
\[
    \frac{(3+\alpha)F}{3F}
    = 1+\frac{\alpha}{3},
\]
excluding the lightweight EMA update and periodic calibration.
For the log-SNR coupling band $[2,\,2.5]$, the saturated active
fraction is $\alpha=0.625$, yielding a multiplier of
approximately $1.208\times$ before calibration.
Calibration probes at most two log-SNR points using 16 batches
per point each epoch, contributing an amortized cost of
approximately $0.009\times$ the baseline training cost.
The resulting estimated multiplier is approximately
$1.22\times$ in the saturated regime and $1.21\times$
averaged over the full training run, during which the active
fraction is initially lower.

Wall-clock profiling uses 8 GPUs, a global batch size of 1024,
and SDPA with bfloat16 precision.
For JiT-L/16, the step time increases from $0.544$ seconds
with teacher evaluation disabled to $0.649$ seconds when
the active fraction saturates, corresponding to a
$1.19\times$ multiplier.
Teacher overhead is lower before saturation, when fewer
samples receive alignment supervision.
For JiT-B/16, the saturated step time is $0.205$ seconds.
A separate paired measurement under matched load gives
$0.71$ seconds without teacher evaluation and $0.83$ seconds
with it, corresponding to a $1.17\times$ multiplier.
At 1,251 steps per epoch, the reported saturated step times
correspond to approximately 4.3 minutes for B/16 and
13.5 minutes for L/16, excluding any calibration time not
already included in the step measurements. The training speed is broadly comparable to that of methods
such as Self-Flow and SRA, while \ourwork{} requires no
additional trainable alignment layers.

\section{Additional Results}
\subsection{System-level Comparison}
\label{app:system-comparison}
Table~\ref{tab:system_comparison} provides a broader comparison with latent- and pixel-space diffusion models. We produce better results than our vanilla JiT baselines across scales and even achieve the same performance (FID=1.82) as JiT-G with our~\ourwork-H which is only 1/2 of the model size. 
\begin{table*}[h]
\centering
\caption{System-level comparison on ImageNet $256\times256$. Parameters include the generator and tokenizer decoder used at inference, excluding training-only components. GFLOPs measure one generator forward pass excluding tokenizer computation. \ourwork improves FID over JiT at each evaluated scale without changing inference computation. The specific REPA encoder used by DeCo is unspecified.}
\label{tab:system_comparison}
\vspace{0.5em}
\small
\setlength{\tabcolsep}{5pt}
\renewcommand{\arraystretch}{1.12}
\arrayrulecolor{tableborder}
\resizebox{\textwidth}{!}{
\begin{tabular}{|l l|r r r|r r|}
\hline
\rowcolor{tableheader}
\textbf{Method} & \textbf{External components} & \textbf{Params (M)} & \textbf{Epochs} & \textbf{GFLOPs} & \textbf{FID $\downarrow$} & \textbf{IS $\uparrow$} \\
\hline

\rowcolor{tablegroup}
\multicolumn{7}{|l|}{\textbf{Latent-space diffusion}} \\
\hline
DiT-XL/2~\citep{peebles2023scalable} & VAE & $675+49$ & 1400 & 238 & 2.27 & 278.2 \\
SiT-XL/2~\citep{ma2024sit} & VAE & $675+49$ & 1400 & 238 & 2.06 & 277.5 \\
SiT-XL/2 + REPA~\citep{yu2024representation} & VAE, DINOv2 & $675+49$ & 200 & 238 & 1.96 & 264.0 \\
SiT-XL/2 + REPA~\citep{yu2024representation} & VAE, DINOv2 & $675+49$ & 800 & 238 & 1.42 & 305.7 \\
\hline

\rowcolor{tablegroup}
\multicolumn{7}{|l|}{\textbf{Pixel-space diffusion: existing systems}} \\
\hline
SiD2, UViT/1~\citep{hoogeboom2025simpler} & None & -- & -- & 1306 & 1.38 & -- \\
PixelFlow-XL/4~\citep{chen2025pixelflow} & None & 677 & 320 & 5818 & 1.98 & 282.1 \\
PixelDiT-XL/16~\citep{yu2026pixeldit} & DINOv2 & 797 & 320 & 311 & 1.61 & 292.7 \\
PixelDiT-XL/16~\citep{yu2026pixeldit} & DINOv2 & 797 & 800 & 311 & 1.54 & 297.0 \\
DeCo-XL/16 (Heun)~\citep{ma2026deco} & REPA encoder & 682 & 600 & -- & 1.69 & 304.0 \\
DeCo-XL/16 (Euler)~\citep{ma2026deco} & REPA encoder & 682 & 800 & -- & 1.62 & 301.0 \\
\hline

\rowcolor{tablegroup}
\multicolumn{7}{|l|}{\textbf{JiT backbone comparisons}} \\
\hline
JiT-B/16~\citep{li2026back} & None & 131 & 600 & 50 & 3.66 & 275.1 \\
\rowcolor{tableresult}
\ourwork-B/16 & None & 131 & 600 & 50 & 3.41 & 271.82 \\
\hline
JiT-L/16~\citep{li2026back} & None & 459 & 600 & 176 & 2.36 & 298.5 \\
\rowcolor{tableresult}
\ourwork-L/16 & None & 459 & 600 & 176 & 2.26 & 284.93 \\
\hline
JiT-H/16~\citep{li2026back} & None & 953 & 600 & 364 & 1.86 & 303.4 \\
PixelREPA-H/16~\citep{shin2026representation} & DINOv2 & 953 & 600 & 364 & 1.81 & 317.2 \\
\rowcolor{tableresult}
\ourwork-H/16 & None & 953 & 600 & 364 & 1.82 & 302.8 \\
\hline
\rowcolor{tablegroup}
\multicolumn{7}{|l|}{\textbf{Larger-backbone reference}} \\
\hline
JiT-G/16~\citep{li2026back} & None & 2B & 600 & 766 & 1.82 & 292.6 \\
\hline
\end{tabular}
}
\end{table*}

\subsection{Additional qualitative results}

\begin{figure}[h]
    \centering
    \includegraphics[width=1.0\linewidth]{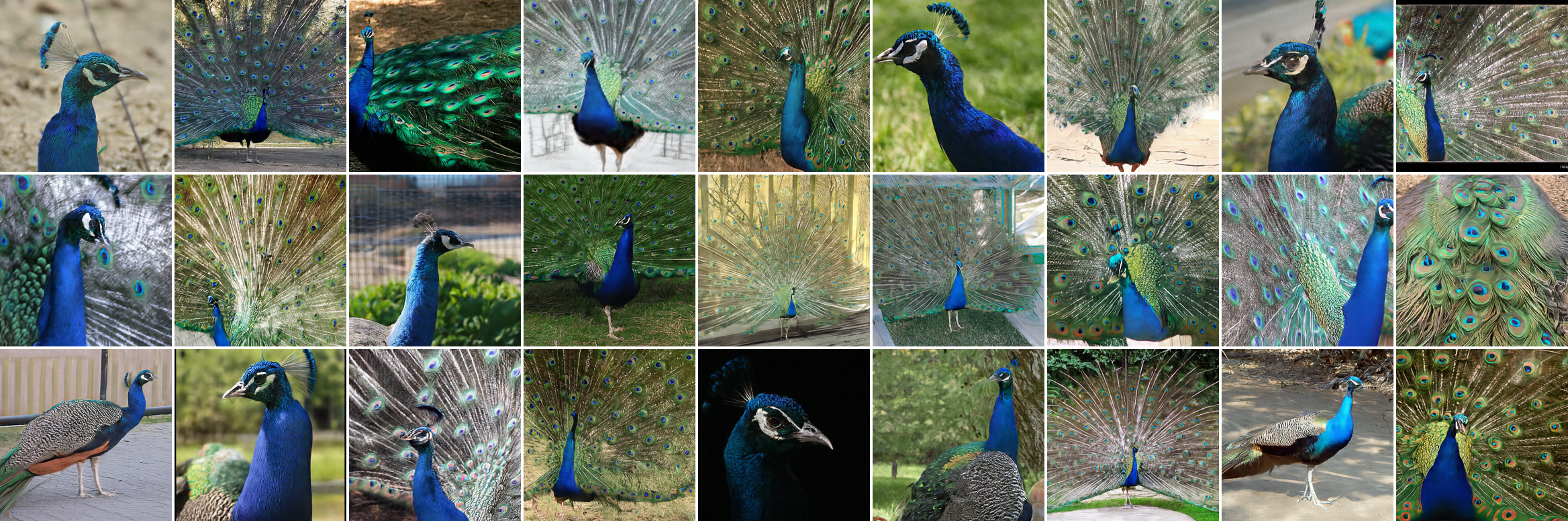}
    \vspace{-20pt}
    \caption{Uncurated samples from Class 084 (peacock)}
    \label{fig:class_084}
\end{figure}

\begin{figure}[h]
    \centering
    \includegraphics[width=1.0\linewidth]{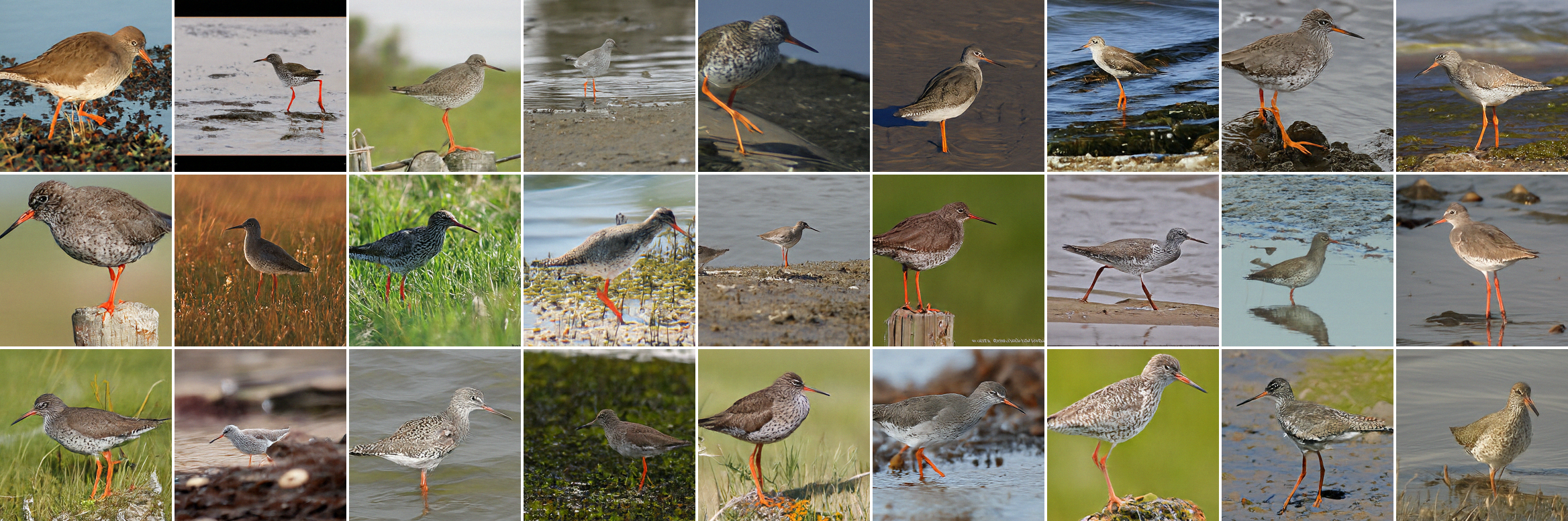}
    \vspace{-20pt}
    \caption{Uncurated samples from Class 141 (redshank)}
    \label{fig:class_141}
\end{figure}

\begin{figure}[h]
    \centering
    \includegraphics[width=1.0\linewidth]{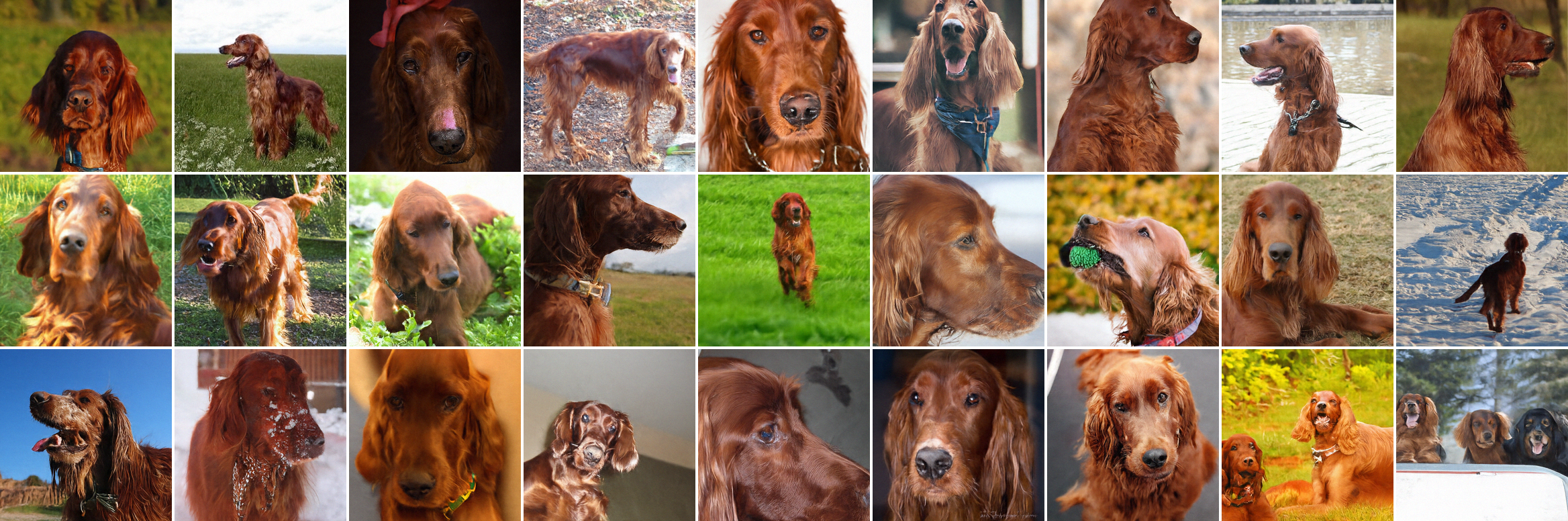}
    \vspace{-20pt}
    \caption{Uncurated samples from Class 213 (Irish setter)}
    \label{fig:class_213}
\end{figure}

\begin{figure}[h]
    \centering
    \includegraphics[width=1.0\linewidth]{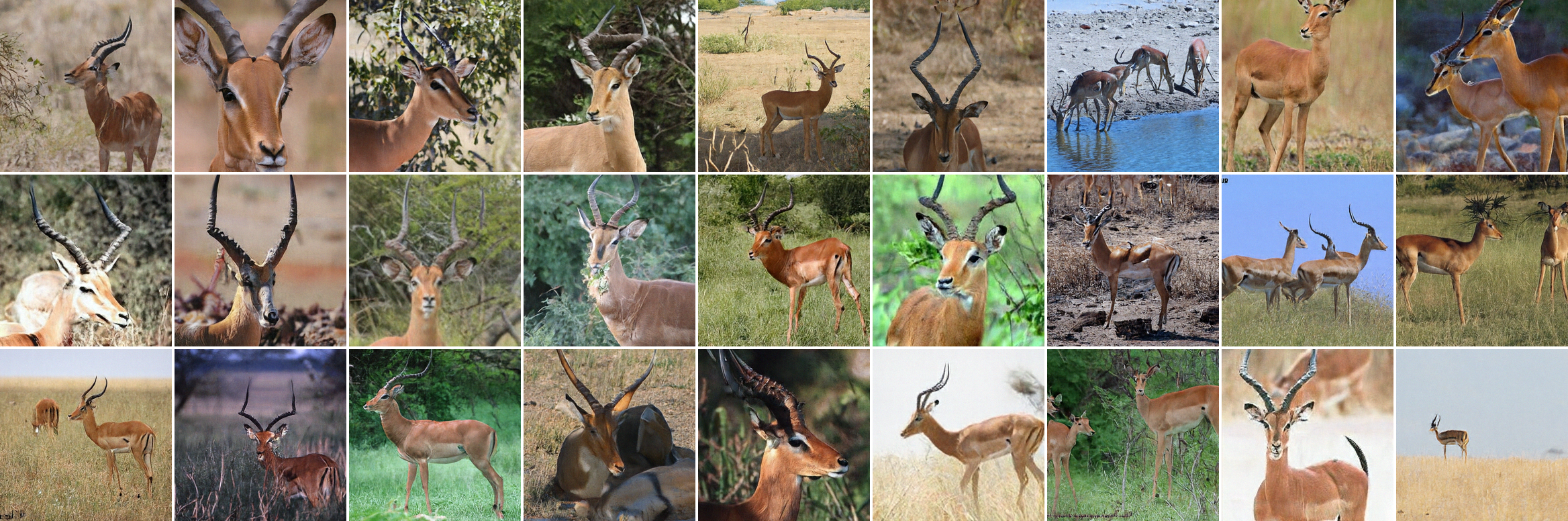}
    \vspace{-20pt}
    \caption{Uncurated samples from Class 352 (impala)}
    \label{fig:class_352}
\end{figure}

\begin{figure}[h]
    \centering
    \includegraphics[width=1.0\linewidth]{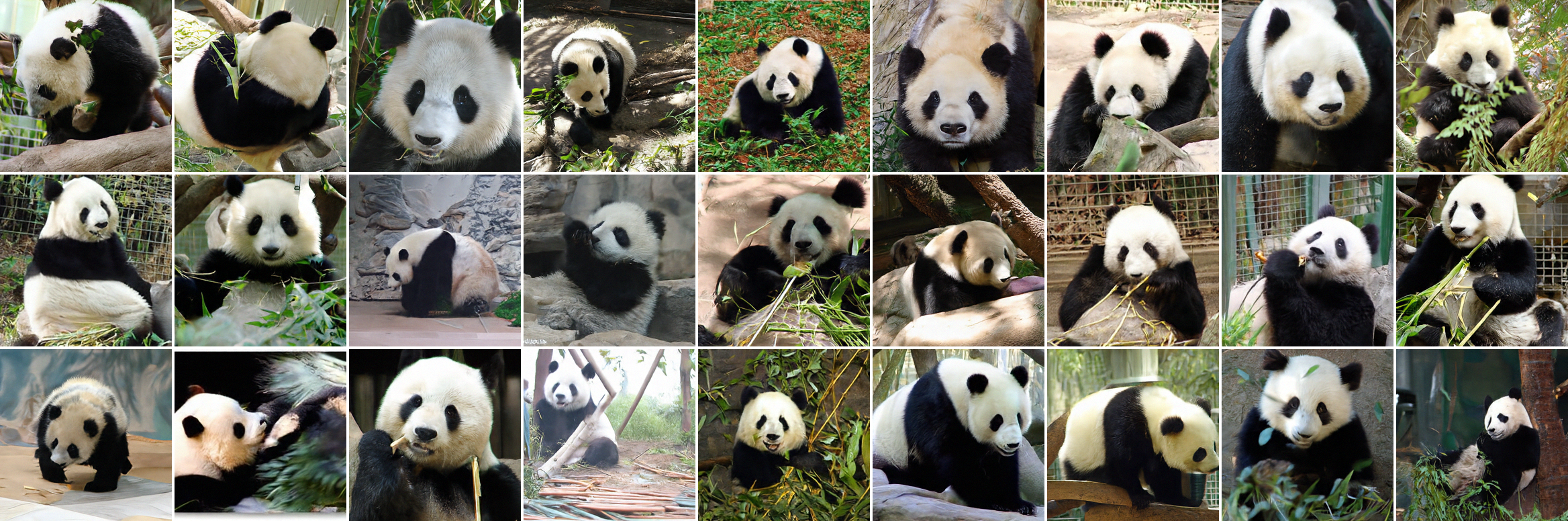}
    \vspace{-20pt}
    \caption{Uncurated samples from Class 388 (giant panda)}
    \label{fig:class_388}
\end{figure}

\begin{figure}[h]
    \centering
    \includegraphics[width=1.0\linewidth]{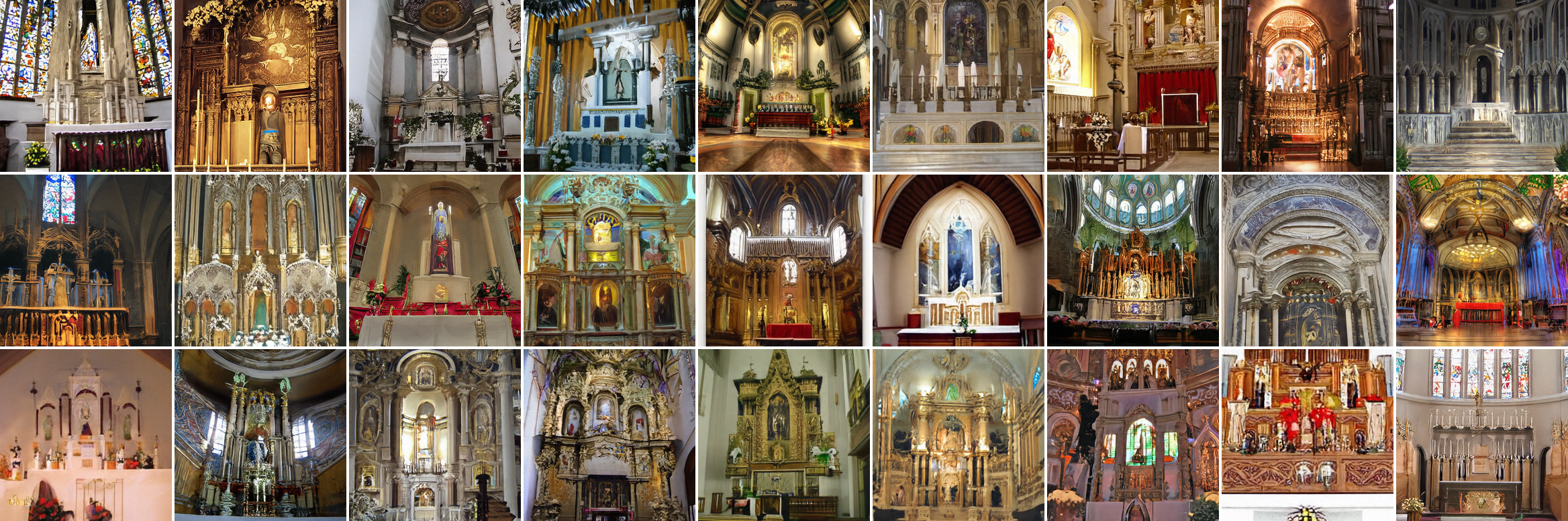}
    \vspace{-20pt}
    \caption{Uncurated samples from Class 406 (altar)}
    \label{fig:class_406}
\end{figure}

\begin{figure}[h]
    \centering
    \includegraphics[width=1.0\linewidth]{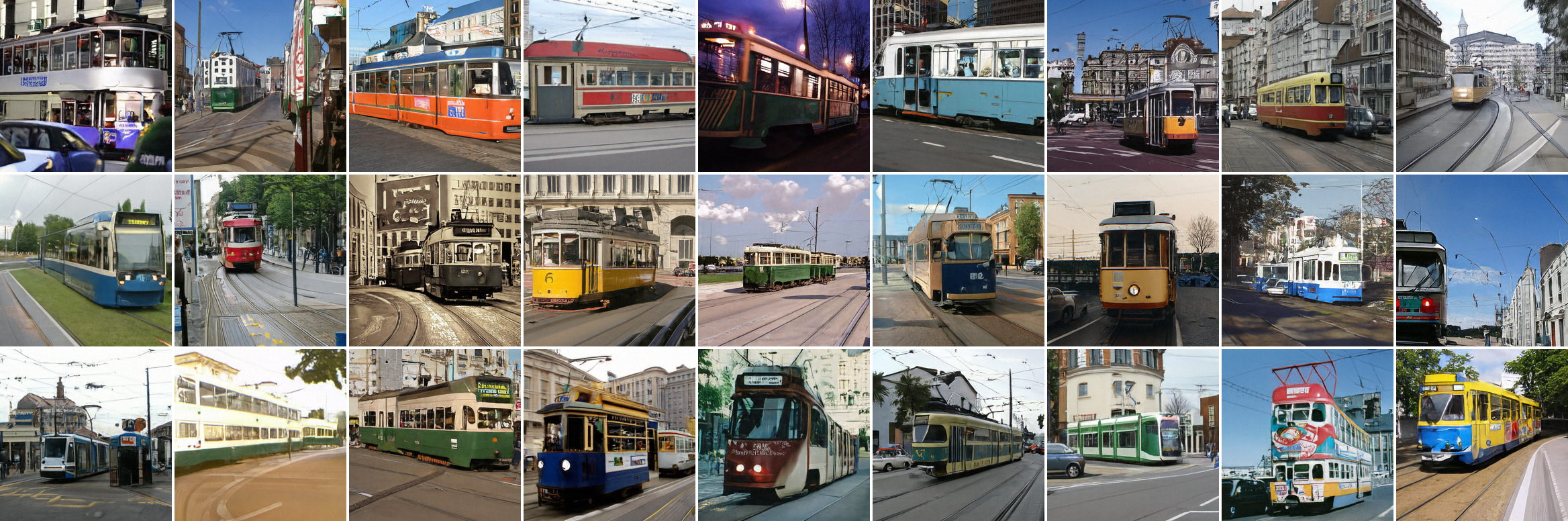}
    \vspace{-20pt}
    \caption{Uncurated samples from Class 829 (streetcar)}
    \label{fig:class_829}
\end{figure}

\begin{figure}[h]
    \centering
    \includegraphics[width=1.0\linewidth]{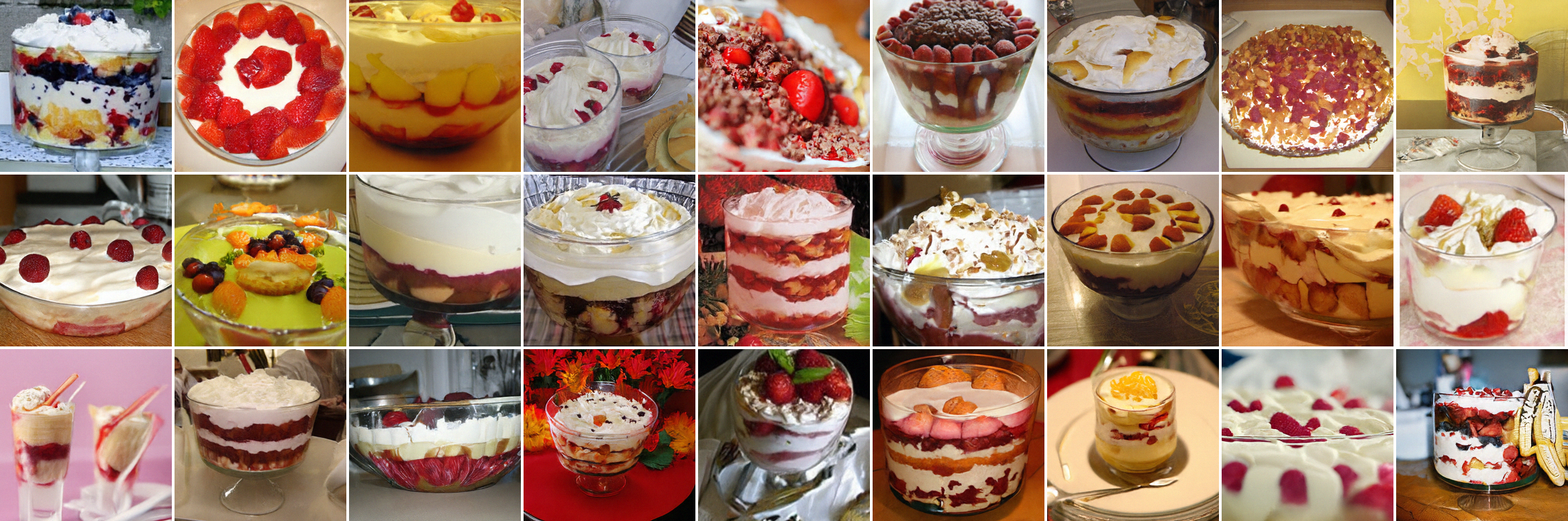}
    \vspace{-20pt}
    \caption{Uncurated samples from Class 927 (trifle)}
    \label{fig:class_927}
\end{figure}

\begin{figure}[h]
    \centering
    \includegraphics[width=1.0\linewidth]{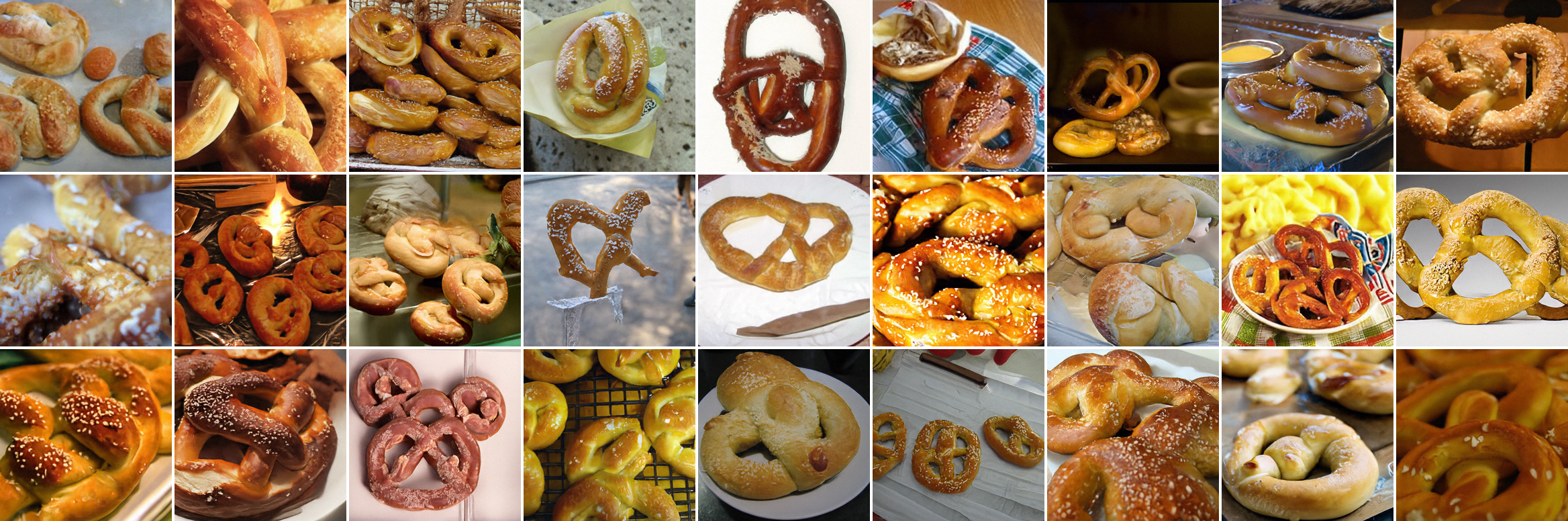}
    \vspace{-20pt}
    \caption{Uncurated samples from Class 932 (pretzel)}
    \label{fig:class_932}
\end{figure}

\begin{figure}[h]
    \centering
    \includegraphics[width=1.0\linewidth]{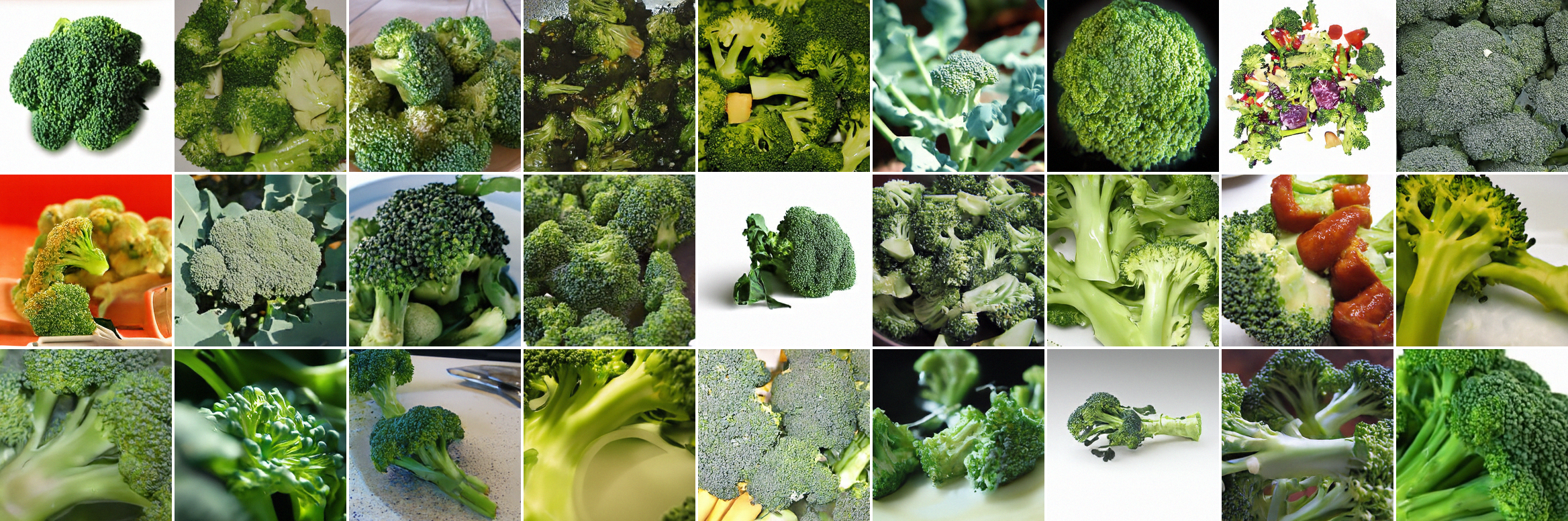}
    \vspace{-20pt}
    \caption{Uncurated samples from Class 937 (broccoli)}
    \label{fig:class_937}
\end{figure}

\begin{figure}[h]
    \centering
    \includegraphics[width=1.0\linewidth]{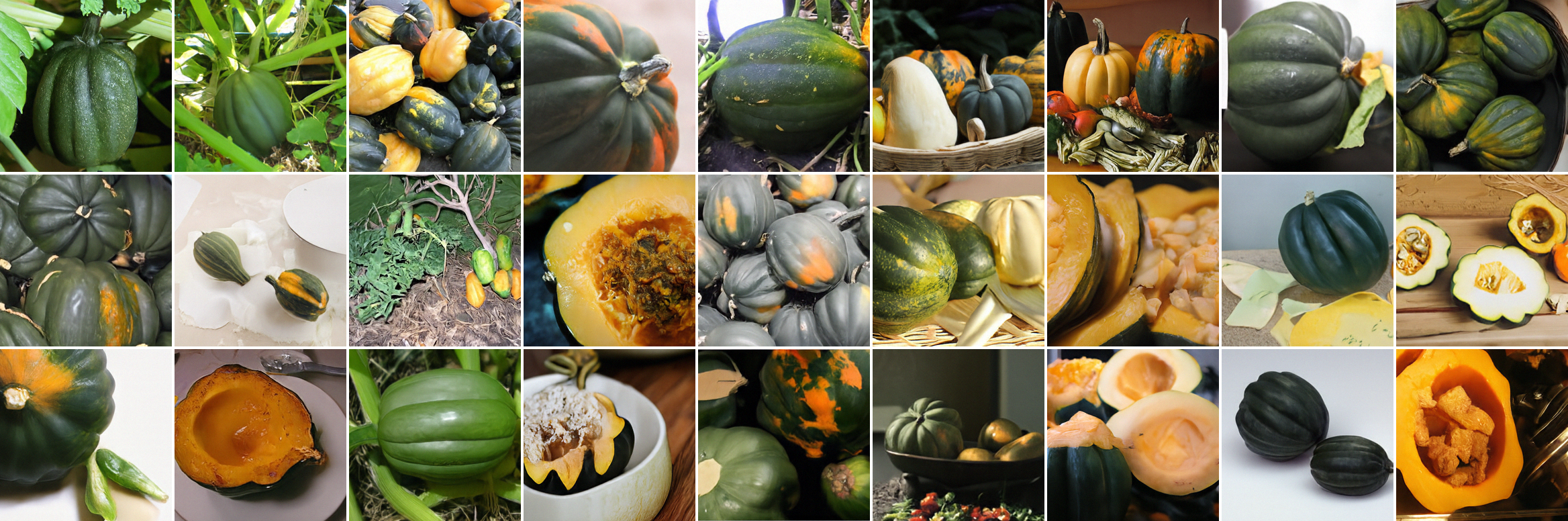}
    \vspace{-20pt}
    \caption{Uncurated samples from Class 941 (acorn squash)}
    \label{fig:class_941}
\end{figure}


\end{document}